\documentclass[letterpaper,12pt,titlepage,oneside,final]{book}
 
\usepackage{ifthen}
\newboolean{PrintVersion}
\setboolean{PrintVersion}{false} 

\usepackage[ampersand]{easylist}
\usepackage{subfig}
\usepackage{algorithm}
\usepackage{algpseudocode}
\usepackage[numbers]{natbib}
\usepackage{setspace}

\ifthenelse{\boolean{PrintVersion}}{
\usepackage[top=1in,bottom=1in,left=0.75in,right=1.25in]{geometry}   
}{
\usepackage[top=1in,bottom=1in,left=0.75in,right=1.25in]{geometry}   
}

\usepackage{amsmath,amssymb,amstext} 
\usepackage{graphicx} 

\usepackage{nomentbl} 
\makenomenclature 

\usepackage{ifpdf}

\newcommand{\href}[1]{#1} 

\usepackage[\ifpdf pdftex,\fi letterpaper=true,pagebackref=false, pdfauthor={Rohan Shrirang Pandey}, pdftitle={Semantic Composition in Visually Grounded Language Models}]{hyperref} 
\ifthenelse{\boolean{PrintVersion}}{   
\hypersetup{	
    citecolor=black,%
    filecolor=black,%
    linkcolor=black,%
    urlcolor=black}
}{} 

\let\origdoublepage\cleardoublepage
\newcommand{\clearemptydoublepage}{%
  \clearpage{\pagestyle{empty}\origdoublepage}}
\let\cleardoublepage\clearemptydoublepage

\begin{document}

%
%
\pagestyle{empty}
\pagenumbering{roman}

\begin{titlepage}
        \begin{center}
        \vspace*{1.0cm}

        \Huge
        {\bf Semantic Composition in Visually Grounded Language Models }

        \vspace*{1.0cm}

        \normalsize
        by \\

        \vspace*{1.0cm}

        \Large
        Rohan Shrirang Pandey \\

        \vspace*{4.0cm}

        \normalsize
        Undergraduate Thesis submitted to the\\
        Language Technologies Institute\\
        in partial fulfillment of the requirements for\\
        Honors from the School of Computer Science

        \vspace*{2.0cm}

        Supervised by:\\
        Dr. Louis-Philippe Morency\\
        Paul Pu Liang

        \vspace*{2.0cm}

        \copyright~Carnegie Mellon University, Pittsburgh, PA, 2023\\
        \end{center}
\end{titlepage}

\pagestyle{plain}
\setcounter{page}{2}

\cleardoublepage 

%
%



\begin{center}\textbf{Abstract}\end{center}

What is sentence meaning and its ideal representation? Much of the expressive power of human language derives from semantic composition, the mind’s ability to represent meaning hierarchically \& relationally over constituents. At the same time, much sentential meaning is outside the text and requires grounding in sensory, motor, and experiential modalities to be adequately learned. Although large language models display considerable compositional ability, recent work shows that visually-grounded language models drastically fail to represent compositional structure. In this thesis, we explore whether \& how models compose visually grounded semantics, and how we might improve their ability to do so. 

Specifically, we introduce 1) WinogroundVQA, a new compositional visual question answering benchmark, 2) Syntactic Neural Module Distillation, a measure of compositional ability in sentence embedding models, 3) Causal Tracing for Image Captioning Models to locate neural representations vital for vision-language composition, 4) Syntactic MeanPool to inject a compositional inductive bias into sentence embeddings, and 5) Cross-modal Attention Congruence Regularization, a self-supervised objective function for vision-language relation alignment. We close by discussing connections of our work to neuroscience, psycholinguistics, formal semantics, and philosophy.

\cleardoublepage


\begin{center}\textbf{Acknowledgements}\end{center}

I would like to begin by thanking my thesis advisors, LP Morency and Paul Liang, for the continual guidance on the several projects this thesis encompasses. The current state of my research would also not be possible without my previous advisors Graham Neubig, Adrian Brasoveanu, Marilyn Walker, Tim Mullen, and Narges Norouzi. I'd also like to thank my collaborators Rulin Shao, Uri Alon, Frank Xu, Kevin Bowden, as well as everyone I've gone to for advice, Tristan Thrush, Chunyuan Li, Tom McCoy, Emmy Liu, Brad Mahon, Brian MacWhinney, Mike Tarr, Chris Potts, and Paul Smolensky.

I'd like to thank my friends Kamil Kisielewicz, Shiv Rustagi, Aryaman Arora, Miguel Tenant de la Tour, Max Alfano-Smith, Adam Farris, Justus Mattern, Bradley Hsu, Roanak Baviskar, Avik Jain, Steven Feng, and Abhinav Tumu for being critical in my journey as a researcher and builder. Also, thank you to the brothers of $\Sigma \Phi$E PA-$\Theta$  for making my time at CMU enjoyable enough to have the energy to write this thesis. I'm also grateful to all the amazing people and teams I worked with at Microsoft (Man Fong), Wordcab (Aleks Smechov), Vizerto (Sanjay Shitole), Bunch (John Wehr), Intheon (Christian Kothe), and SapientX (David Colleen) over the course of the last 4 years.

To my father I am thankful for the pragmatic, solution-seeking mindset I was raised with and to my mother I am thankful for fostering my curiosity and connections to the Indian intellectual tradition, without which much of my motivation on these questions would be absent. In that vein, I thank my intellectual forefathers in whose writings I found meaning in semantics: Bharadv\=aja B\=arhaspatya, P\=a\d{n}i\d{n}i, K\=aty\=ayana, Jaimini, Bhart\d{r}hari, Kum\=arila Bha\d{t}\d{t}a, V\=acaspati Mi\'sra, Ga$\dot{\text{n}}$ge\'sa Up\=adhy\=aya, and Vimalak\d{r}\d{s}\d{n}a Matilal.

Finally, I thank the reviewers of all work included in this thesis for their insightful and valuable comments.

\cleardoublepage





\begin{figure}
    \centering
    \vspace{6cm}
    \includegraphics[width=0.8\linewidth]{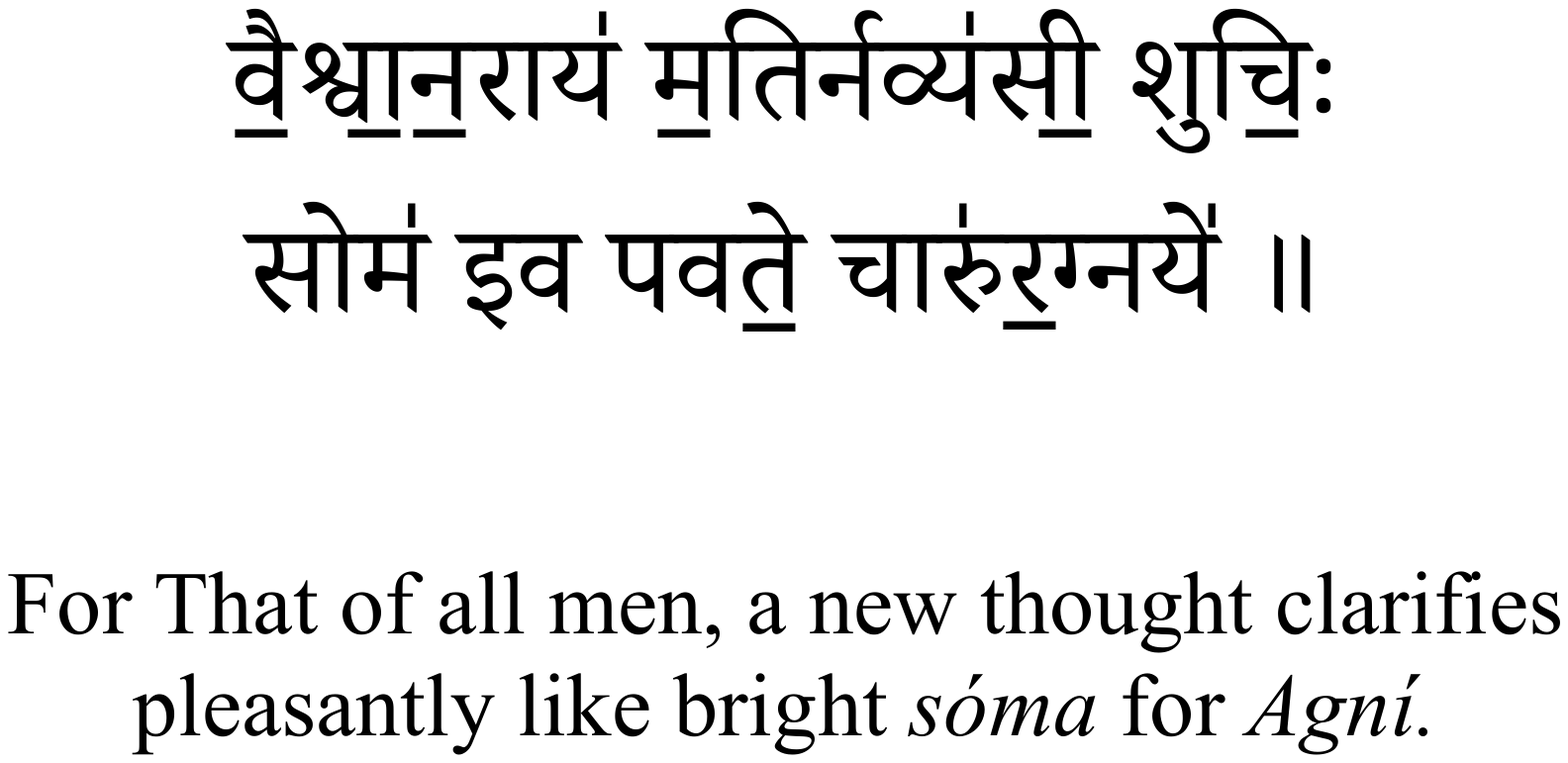}
\end{figure}

\cleardoublepage


\renewcommand\contentsname{Table of Contents}
\tableofcontents
\cleardoublepage
\phantomsection




%

\pagenumbering{arabic}

\chapter{Introduction}

The principle of compositionality and problem of symbol grounding have a long history in Cognitive Science \& Analytic Philosophy, tracing back to Gottlob Frege in the late 19th century; an even longer history may be found among the Ny\=aya \& M\={\i}m\=a\.ms\=a schools c. 7-12th century \cite{matilal70}. For much of the 20th century, logicians \cite{russell}, formal semanticists \cite{montague}, and theoretical syntacticians \cite{chomsky} explored how symbolic rule systems could describe the compositionality of human language without trying to describe how the symbols might map onto the multimodal world our language interacts with.

\section{Semantic Composition}

At its core, compositionality is about the regularity of a system. In a fully compositional system, the meaning of an expression is perfectly predictable by its constituents and their relational structure \cite{partee2008compositionality}. Compositionality is a useful property of language because it enables high productivity and generalization while requiring less representational capacity \cite{baroni2020linguistic}.

Although strictly compositionality is a measurement of a linguistic system, we often discuss a model's compositional ability or understanding. By this, we mean a model builds representations that capture the underlying compositional structure in its input. Neither humans nor language models are explicitly trained on syntax trees, but humans implicitly learn syntax which enables them to understand compositional distinctions in semantics.

Compositional ability is vital to forming structured representations, symbolic reasoning, and various high-level system 2 cognitive capacities \cite{bengio2019system}. Ultimately, semantics is about interpretation---assigning a meaning to an expression abstracted away from the specifics of a language. Interpretation is what allows us to bridge the gap from one language to another language or modality. Traditionally, this meant interpreting utterances like ``the car is red" as true if and only if the car is red, locking us into a self-referential system with symbols all the way down.

\section{Visual Grounding}
Although there is certainly utility in approaches that elegantly lift the compositional structure of language to a more formalized level, one cannot deny the dissatisfaction of purely symbolic semantic interpretations. How can we know if the car is actually red when we can't look up from the page? How can we ground our semantic interpretation in something beyond symbols, as humans do through interaction with the world?

The problem of symbol grounding is quite broad, with the earliest Western ideas exploring the relationship between sign \& meaning \cite{saussure1916} and sense \& reference \cite{frege1892}. Simply put, a distinction must be made between the surface level linguistic expression, the meaning it conveys, and the entities in the world that it refers to. While considerable progress has been made over the past century formalizing the relation between the first 2 (syntax \& semantics) \cite{montague1970}, relatively little progress had been made on grounding semantics to sensory representations of the world until the deep learning era.

Prior to deep learning, it was intractable to compute high quality image representations, relegating all progress in semantics to more symbolic domains. In the deep learning era, the shared semantic framework of vectorial embeddings enables interaction of language meaning representations with image representations. Grounding a model's semantics in sensory modalities will be critical for developing robots that can see, hear, and feel the real world \cite{bisk2020experience}.

\section{Observations}

In order to better understand the problem we face---visually grounding the compositional semantics of language---let us begin with a Wittgensteinian (or perhaps s\=utranic), first principles approach to observing some general truths about the vision-language duality.

\begin{easylist}[tractatus]
\ListProperties(Margin1=0cm, Progressive=.5cm)

& Fundamentally, both visual (2D) and audio-linguistic (1D) stimuli are unstructured.
& The ability to identify constituents (objects or tokens) of a stimulus (scene or sentence) is learned with inductive biases.
&& Contiguity 
&&& An object usually occupies a contiguous region of space.
&&& A token usually occupies a contiguous span of time.
&& Repetition
&&& Seeing a cluster of pixels corresponding to a rabbit in multiple scenes.
&&& Hearing a sequence of syllables corresponding to “bunny” in multiple sentences.
& A constituent is usually an instance of a category (class or word).
&& A set of objects from different scenes can visually vary considerably but still instantiate the same class.
&&& A leaf may appear with different colors, occlusions, orientations depending on context.
&& A set of tokens from different sentences can auditorily vary considerably but still instantiate the same word
&&& "Leaf" may be heard with different phonetics, prosody, or morphology depending on context.
& Every visual class corresponds to a word.
& Some words correspond to a visual class.
& Constituents form relations with each other in a stimulus.
&& Objects are organized in visual space.
&&& person ON bed; blanket ON person; eyelids IN person; top-eyelids TOUCH bottom-eyelids.
&& Tokens are ordered in time.
&&& “The man sleeps”.
& Constituents in a stimulus are usually modified for relations to form.
&& A blanket ON a sleeping person will not be neatly folded or completely flat, it will broadly contour their body.
&& “sleep” is modified to “sleeps” when related to a single subject.
& There are restrictions on valid stimuli.
&& Two objects in a scene may not occupy the same spatial position.
&& Two tokens in a sentence may not fill the same argument of a relation.
& A scene may be described by different sentences.
& A sentence may be interpreted as different scenes.

\end{easylist}

\section{Vision-Language Deep Learning}


Vision-language deep learning research \cite{liang2022foundations} synthesizes advances in computer vision and natural language processing to build models capable of solving multimodal tasks like image-text matching, image2text captioning, and text2image generation. There are a number of methods for learning joint representations of vision and language, but most leverage the transformer architecture \cite{vaswani17}. 

Encoder models generally either use a dual unimodal encoder architecture \cite{radford2021clip}, a single cross-modal encoder \cite{chen2020uniter}, or a combination \cite{singh2022flava}. Image-conditioned text generation models often consist of an image encoder that feeds into a causal transformer decoder \cite{li2022blip}. Text-based image generation models are quite varied but often condition an image diffusion model on the output of a text encoder \cite{ramesh2021dalle2}.

\subsection{Problem}

While several results in vision-language deep learning are quite impressive, recent work has shown that models consistently run into issues capturing compositional structure. Winoground \cite{thrush22} is a simple vision-language compositionality task that tests a VLM's ability to match syntactic permutations of a caption to their corresponding images, finding that all recent and state-of-the-art VLMs perform below chance levels on the Winoground evaluation dataset. Contemporaneously, \citet{milewski22} probe for structural knowledge in VLM's, finding that they encode significantly less linguistic syntax than LM's and virtually no visual structure. \citet{tejankar2021fistful} find that VLM's perform no better than a bag-of-words at an image-text matching task. 

In order to improve these abilities, \citet{nikolaus2022vision} find that VLM's trained using fine-grained vision-language objectives display better syntactic understanding. \citet{yuksekgonul2022bags} show that training VLM's with order-permuted hard negatives improves compositional performance. We seek to contribute to contribute to this pool of work that addresses issues of fine-grained, compositional, or syntactic understanding in visually grounded language models.


\section{Outline}

In this thesis, we explore whether \& how models compose visually grounded semantics, and how we might improve their ability to do so.

First, we propose WinogroundVQA (Sec. \ref{sec:wgvqa} \footnote{Results in progress}) to test the ability of generative vision-language foundation models to capture compositional distinctions, building on previous work that only tested image-text matching models.

Next, we propose methods for understanding how models perform semantic composition. Syntactic Neural Module Distillation (Sec. \ref{sec:synnamon}) enables us to test whether a sentence's syntax tree is a strong causal model of its embedding's composition. We build a simple visualization tool (Sec. \ref{sec:attflow}) to observe if a transformer's attention activation flow, seen as its causal structure, aligns with a sentence's syntax, a symbolic model of its composition. We then apply this intuition more rigorously to adapt a mechanistic interpretability approach, causal tracing, to locate neural representations important for composition in image captioning models (Sec. \ref{sec:vlmechinterp} \footnote{Experiment implementation in progress}).

Having better understood composition in visually grounded models, we now explore approaches for improving their ability. We propose CACR (Sec. \ref{sec:cacr}), a self-supervised objective that encourages models to represent unimodal relations in a way that better enables cross-modal relation alignment, improving on the current Winoground state-of-the-art. To inject a compositional inductive bias into unimodal embeddings, we propose the Syntactic MeanPool (Sec. \ref{sec:synavg}) which seeks to improve CLIP's image-text match score on Winoground. 

In closing, we draw connections to the cognitive sciences (Sec. \ref{ch:conclusion}). We explore how some of the behaviors we observe in models may have parallels in the brain since multimodal semantic representations are localized in the left anterior temporal lobe. We discuss vision-language compositional ability from a psycholinguistic perspective by testing a time-constrained version of Winoground on humans. We finally conjecture a multimodal semantics framework to formalize compositional behavior in vision-language embedding space that draws inspiration from homotopy type theory. 


\chapter{Can models compose visually grounded semantics?}

\section{Past Work}

Several datasets to test vision-language models' compositional ability have recently cropped up \cite{thrush2022winoground, ma2022crepe, zerroug2022cvr}, but the question has been studied since early in the deep learning era \cite{suhr2017nlvr, antol2015vqa}. We're particularly interested in Winoground \cite{thrush2022winoground} and how we can leverage its general approach to benchmarking visually grounded compositional ability. The Winoground dataset consists of pairs of image-caption pairs, where the captions share the same words but in a slightly different order---for example, ``mug in grass'' v.s. ``grass in mug''. If a vision-language encoder model produces a higher image-text match score between the correct pairs than the incorrect pairs, then it understands the compositional distinctions.

Just as unimodal NLP research has shifted from encoder models like BERT \cite{devlin-etal-2019-bert} to generative models like GPT-3 \cite{brown2020language} over the past few years, vision-language research appears to be shifting now from joint encoder models like UNITER \cite{chen2020uniter} and FLAVA \cite{singh2022flava} to image-conditioned text generation models like BLIP2 \cite{li2023blip2} and GPT-4 \cite{openai2023gpt4}. Unfortunately, it is not straightforward to evaluate these models on Winoground. One could extract the conditional probability of a caption given an image, but these scores are often unavailable due to large models being hosted on private servers with restricted access.

Furthermore, recent work \cite{diwan2022winoground} shows that 19.5\% of Winoground examples require complex reasoning over the image and text. Capturing the complexities of a caption's reasoning in a single embedding and representing an image in such a way that it can be reasoned over using a simple cosine similarity seems difficult and perhaps topologically implausible, making it quite unlikely that two stream models trained using a cross-modal contrastive loss would be able to solve these examples. 

Empirically, this is partially backed by FLAVA's \cite{singh2022flava} performance on Winoground, which is only 9.00 when measured contrastively but 14.25 when measured using image-text matching score. In this latter setting, the model is given the ability to attend cross-modally between vision and language, potentially enabling a simple form of reasoning in the forward pass. By switching to a generative setting, we enable the model to perform deeper reasoning, opening the possibility for visually-grounded chain-of-thought prompting \cite{wei2022chain} on our task.

Another concern raised by \citet{diwan2022winoground} is that 12.5\% of Winoground examples contain unusual text which is out of distribution from standard image captioning datasets or even borderline ungrammatical.

\section{WinogroundVQA Benchmark}
\label{sec:wgvqa}


To address these issues, we propose WinogroundVQA, the Winoground dataset rephrased as a visual question answering problem. Instead of presenting an image-text pair to an encoder model and calculating the match score, we provide an image along with a question like ``Is the mug in the grass?" and score the model based on whether it answers ``yes'' or ``no''. This approach solves a number of the previously mentioned issues:

\begin{enumerate}
    \item It enables us to test the Winoground ability of visually grounded large language models that don't produce a match score and don't expose caption probabilities.
    \item It places Winoground in a more cognitively plausible setting where reasoning is handled through language generation rather than in a single embedding.
    \item It cleans up out of distribution captions by converting them into questions that are phrased less unusually.
\end{enumerate}

Methodologically, we convert Winoground captions to our VQA form with a simple prompt that we query GPT-3.5 with:
``\texttt{Here is a description of a picture: \{caption\}. Rewrite the description as a yes-or-no question but change as little of the phrasing as possible.}''


\subsection{Results}

\begin{table}[h]
    \centering
    \begin{tabular}{c|c c c}
         \textbf{Model} & \textbf{Text} & \textbf{Image} & \textbf{Group}  \\
         \hline
         BLIP & 2.0 & 2.25 & 1.5 \\
         MiniGPT-4 & 6.5 & 6.25 & 5.75 \\
    \end{tabular}
    \caption{Performance of some visually-grounded language models on WinogroundVQA}
    \label{tab:wgvqa_results}
\end{table}
\chapter{How may models compose visually grounded semantics?}

\section{Past Work}
The principle of semantic compositionality suggests that the meaning of a sentence should derive from its subconstituents in a regular, structured fashion \cite{montague1970english}. In recent years, transformers \cite{vaswani17} have become effective at producing sentential meaning representations useful for downstream tasks such as Natural Language Inference, Image-Text Matching, and Document Classification \cite{conneau17, radford21}. However, it has famously been conjectured that "You can't cram the meaning of a whole \%\&!\$\# sentence into a single \$\&!\#* vector" \cite{mooney14}. Since recent models do appear to capture sentence meaning effectively, one wonders how they compose arbitrarily many word meanings together such that their relational structure is captured in a single, fixed-dimensional sentence embedding.

Much work has sought to probe these models for syntax representations and their causal relevance to embedding output. \citet{conneau18} train linear probes to determine if models encode syntactic features like tree distance and depth \cite{krasnowska19, hewitt19}. One line seeks out direct mappings between neural representations and tree structures \cite{mccoy18, chrupala19, jawahar-etal-2019-bert, soulos20, murty2023characterizing}. Other work raises methodological issues with probing \cite{eger19, zhu-rudzicz-2020-information} such as choice of formalism \cite{kuznetsov20} and semantic entanglement \cite{maudslay21}. \citet{ravichander21} raise the possibility that probing may identify causally un-used features; \citet{tucker21} partly address this concern to show that some syntactic features are causally relevant. Another line of work explores the geometry of semantic representations \cite{reif19, hernandez21} and the linearity \cite{baranvcikova19} of syntactic analogies \cite{zhu20}.

Rather than directly analyze sentence embedding models, Neural Module Nets \cite{andreas16} seek to improve compositionality by modularizing semantic functions. We see this effort as ultimately similar to probing for structural representations since the former explores whether explicit structure improves performance and the latter explores whether performant models implicitly learn structure. \citet{geiger21} discovers logical tree causal structures in BERT and \citet{wu21} then guides model distillation using this structure.

\section{Syntactic Neural Module Distillation to Probe Compositionality}
\label{sec:synnamon}

\begin{figure}[h]
    \centering
    \includegraphics[width=3in]{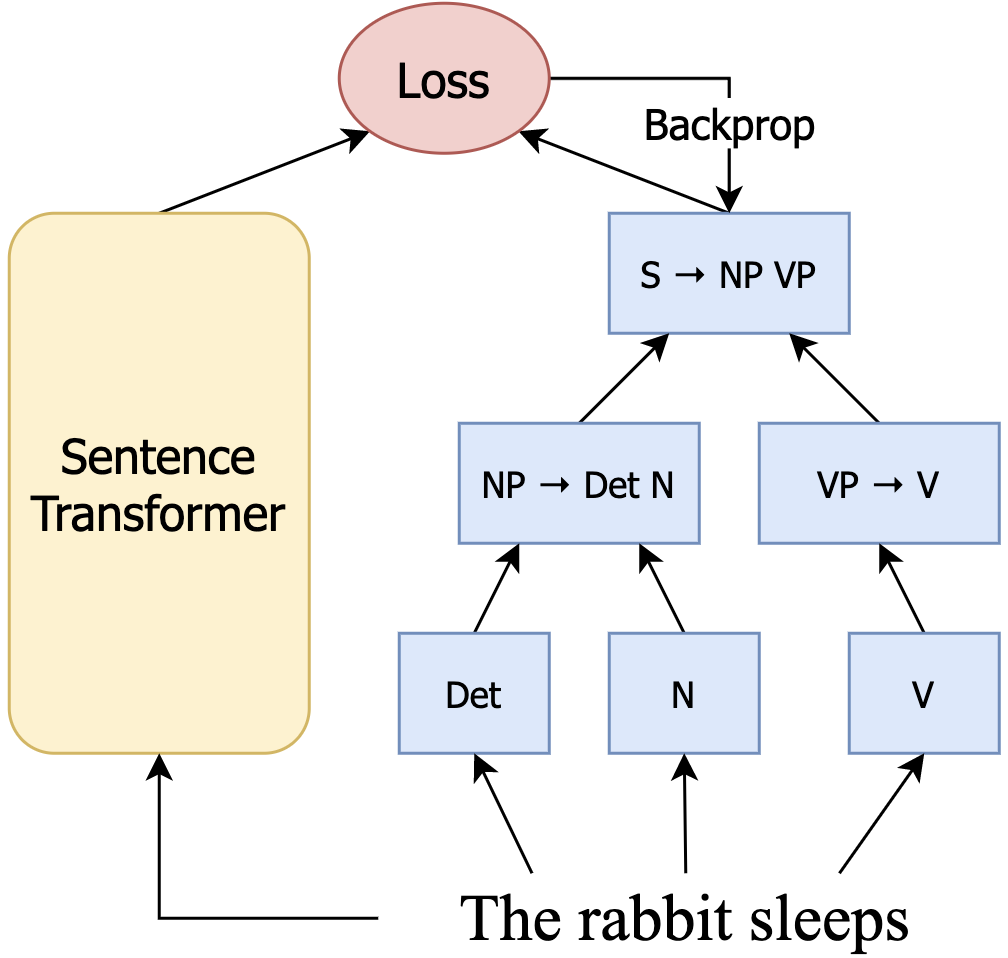}
    \caption{Distilling a transformer to a neural module net structured by the sentence's syntax}
    \label{fig:distillation}
\end{figure}

Our work builds on these findings by strictly taking syntax as the causal structure of sentential semantics and linearity as the geometry of syntax-guided composition; we conduct experiments to test the distillability of transformer-based sentence embedding models to a Syntactic NeurAl Module Net (SynNaMoN), an architecture we introduce that implements these two priors. The extent to which a model can be distilled to a SynNaMoN tells us about its internal syntax representations \& compositional ability.

\subsection{Methods}

\begin{figure}[h]
    \centering
    \includegraphics[width=3in]{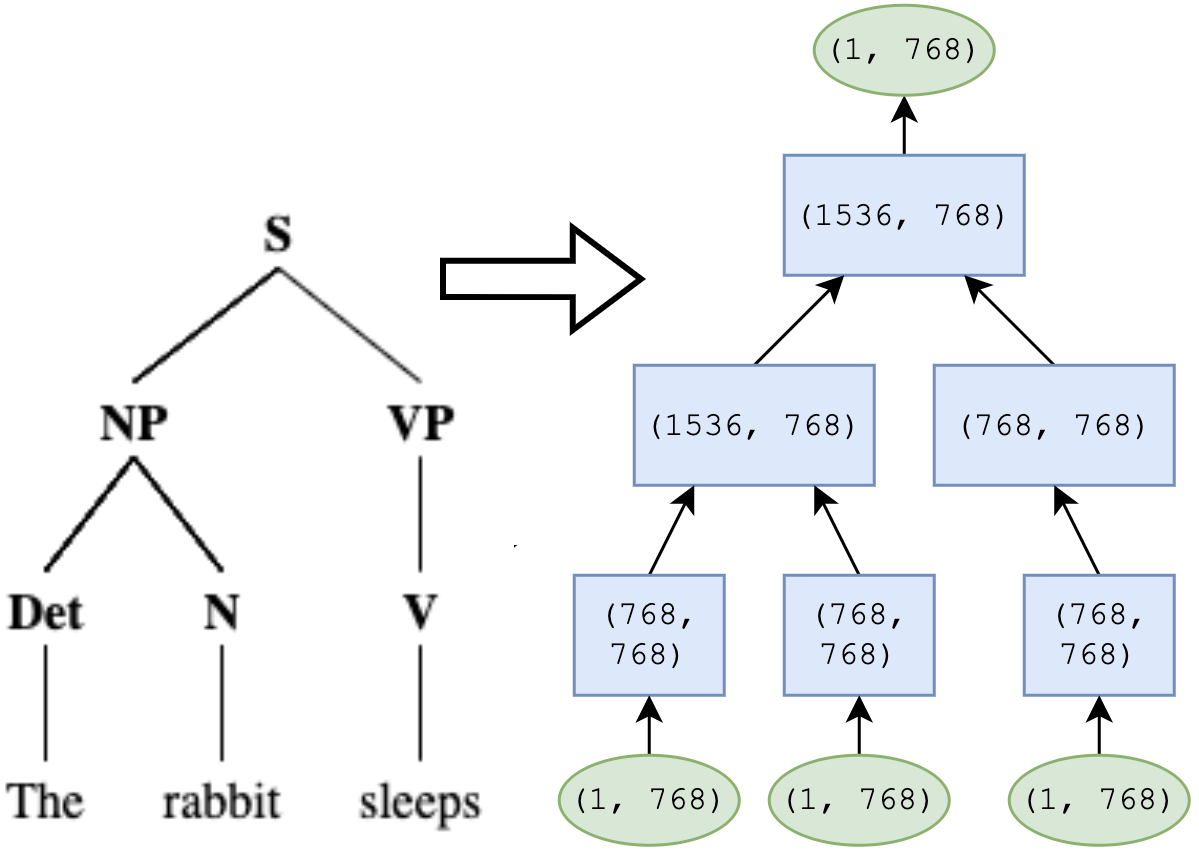}
    \caption{Constructing a sentence's SynNaMoN from its syntax tree; module input and output dimensionalities labeled on the right.}
    \label{fig:syn2mod}
\end{figure}

Unlike prior work \cite{andreas16qa, cirik18}, SynNaMoN modules don't approximate high-level objectives like `Find' or `Count' but rather correspond to specific syntactic rules like `S $\rightarrow$ NP VP' and `NP $\rightarrow$ DT JJ NN'. Each module receives an input of dimensionality $(1, N*D)$ where $N$ is the number of constituents on the syntax rule's right-hand side, and D is the dimensionality of the embedding space (768)—in other words, the input embeddings are concatenated. Though computationally more expensive, concatenation enables the module to learn an arbitrary function over the inputs rather than restricting it to a function over their sum or mean; this enables the module to converge on its ideal composition function which is likely not invariant under summation or averaging. Finally, though our implementation of SynNaMon includes `part-of-speech' modules at the bottom of the parse tree, one could conceivably remove this bottom layer with the hypothesis that the word embeddings already capture part-of-speech information.


\subsubsection{Internal Module Architecture}
\label{sec:internalarch}
To explore the geometry of semantic composition under syntax, we implement 3 module architectures: a linear layer (\textbf{Linear}), a linear layer + a ReLU activation (\textbf{Nonlin}), and a linear layer + ReLU + another linear layer (\textbf{Double}). We explore these 3 architectures to see whether syntax is enough of an inductive bias to linearly approximate sentence embeddings, or if adding non-linearities and additional layers considerably improves performance. The extent to which adding parameters improves our approximation of the teacher model beyond the syntactic structure alone could reveal how much isn't captured by this inductive bias.

\subsubsection{Linguistic Formalism}
We choose to use the Transformational Grammar presented by Penn Treebank \cite{marcus94}, but in principle any Constituency Grammar could be easily used with SynNaMoN, and Dependency Grammars can be adapted with some effort. Since prior work has shown how the choice of linguistic formalism can significantly influence probing results \cite{kuznetsov20}, we float the possibility of such an effect being at play in this work as well. If a student SynNaMoN fails to capture much of the teacher embedding model, perhaps it isn't because of the teacher's non-compositional causal structure, but rather because the formalism used to structure the SynNaMoN is inadequate. Indeed, recent state-of-the-art neural approaches to syntax parsing have learned grammatical tagsets that often differ starkly from human-produced syntactic theories \cite{kitaev2022learned}. We leave these problems to future work which may explore the exciting possibility that certain linguistic formalisms (perhaps even semantic rather than syntactic) are better proxies for a model's compositional structure than others.

\subsection{Experiments}

\label{sec:experiments}
Our main experiment runs 5 sentence embedding models (BERT-base \cite{devlin-etal-2019-bert}, MP-Net \cite{song2020mpnet}, GTR-T5-base, GTR-T5-large, and GTR-T5-xl \cite{ni2021gtr}) on 3 SynNaMoNs with differing internal architectures (Linear, Nonlin, Double; see Sec. \ref{sec:internalarch}). For BERT-base, we extract input word embeddings for each token and use the CLS token as the sentence embedding as is common practice. For the other 4 models, we encode each token alone to serve as its embedding and use the output as the sentence embedding. When words are encoded as more than 1 token, we compute the mean across the subtokens to serve as its word embedding.

In order to heuristically select a learning rate, 5 training runs were conducted with SynNaMoNs optimizing for BERT-base, and learning rate manually set at increments between $10^{-5}$ and $10^{-3}$. We finally chose a rate of $5\times10^{-5}$, but recognize from results that optimal learning rate will likely vary by teacher model \& SynNaMoN internal architecture. 
Analysis would best be reported on the optimal scores achieved by a SynNaMoN after hyperparameter tuning, but due to compute restrictions (1 NVIDIA K80 GPU with 12GB of RAM), this was unfeasible.

Additionally, due to the number of modules (originally ~900, each with ~1M parameters on average), we encountered frequent out-of-memory errors both on CPU \& GPU. Since each module corresponds to a syntax rule and is initialized upon encountering the rule in the dataset, we constrained our data to minimize the number of modules needed.

Specifically, we first constrained our trees to those of height 4 \& 5 (n=16492) in PTB, and then further constrained the trees to those that use a subset of the 300 most common production rules among them. This resulted in 1494 trees, from which we generated a train-validation split of 1250-244. Furthermore, we ensured that all the productions present in trees of the validation split were also included amongst trees in the training split. All this finally resulted in 273 production rules present in our dataset, and the instantiation of 273 modules.

\subsection{Results}
In Tab. \ref{tab:val_loss}, we present scores for all 5 sentence embedding models across the 3 SynNaMon architectures. We compute the average MSE between sentence embeddings in the complete dataset for each model and divide each model's MSE loss by this mean distance to normalize results. The normalized scores we present may intuitively be seen as the portion of variance in a model's sentence embeddings that a SynNaMoN fails to explain. From a probing perspective, the lower a model's score, the more it can be causally approximated by composition along syntactic lines.


\begin{table}[h]
    \centering
    \begin{tabular}{c|c|c|c}
         Sent. Emb. Model & Linear & Nonlin & Double  \\
         \hline
         BERT-base-CLS & .765 & 4.17 & .625 \\
         MP-Net-base & .606 & .963 & .538 \\
         GTR-T5-base & .541 & .844 & .499 \\
         GTR-T5-large & .550 & .898 & .502 \\
         GTR-T5-xl & \textbf{.536} & \textbf{.775} & \textbf{.498} \\
    \end{tabular}
    \caption{Best validation MSE loss of sentence embedding models on each SynNaMoN probe, normalized by chance-level MSE between embeddings}
    \label{tab:val_loss}
\end{table}

First, notice that GTR-T5-xl outperforms all the other models across all the SynNaMoN architectures. This seems to confirm our expectation that larger models should produce more compositional sentence embeddings. However, GTR-T5-xl only marginally outperforms other sizes of GTR-T5 (except on Nonlin, for which it does far better), suggesting that size actually isn't a significant factor in compositionality. The lower performance of GTR-T5-large further corroborates this, but considering its anomalously lower average embedding MSE, the issue requires more work. The fact that GTR-T5 models all display high compositionality despite variance in size suggests something about their architecture or training approach is important—perhaps the representational bottleneck.

All 3 GTR-T5 models perform better than MP-Net, which in turn outperforms BERT CLS. This first fact is slightly surprising considering that on standard sentence representation tasks \cite{reimers19}, MP-Net (63.30) marginally outperforms all GTR-T5 (base: 59.40, large: 62.38, xl: 62.88) models. Evaluation of these sentence embedding models on large-scale, human-interpretable compositionality tasks may reveal that GTR-T5 does indeed produce better compositional representations than MP-Net. Although BERT's CLS token embedding is widely used for sentence representation, these results show that it fails to capture nearly as much compositional information as more targeted sentence embedding models.

Next, observe that although distilling to a Double SynNaMoN is intuitively easier than to a Linear SynNaMoN due to increased parameterization, there aren't always major improvements in distillability. It is possible that the geometric expressivity of the Double SynNaMoN will kick in with scaling of training data, but we hypothesize that this Double score will still approach a limit for all sentence embedding models. This is because syntax only describes a subset of sentence meaning, and the strictness of SynNaMoN's structure prevents this non-compositional component from being learned.

For example, a strictly syntactic compositional interpretation of "village on the river", would represent the village as being literally on top of the river since this is the semantic geometry learned for syntactic structures of the form "NP on NP". A SynNaMoN that includes non-linearities may better learn the geometry of this non-literal "on" relation, but a transformer model would best learn to handle non-compositional phrases due to its lack of strict syntactic constraints. Our broader takeaway from comparing Linear \& Double scores is that much composition along syntactic lines is linear, and non-linearities in transformers primarily serve a purpose other than syntax-guided composition—perhaps in handling non-compositional phrases.


\begin{figure}[h]
\centering
\includegraphics[width=3in]{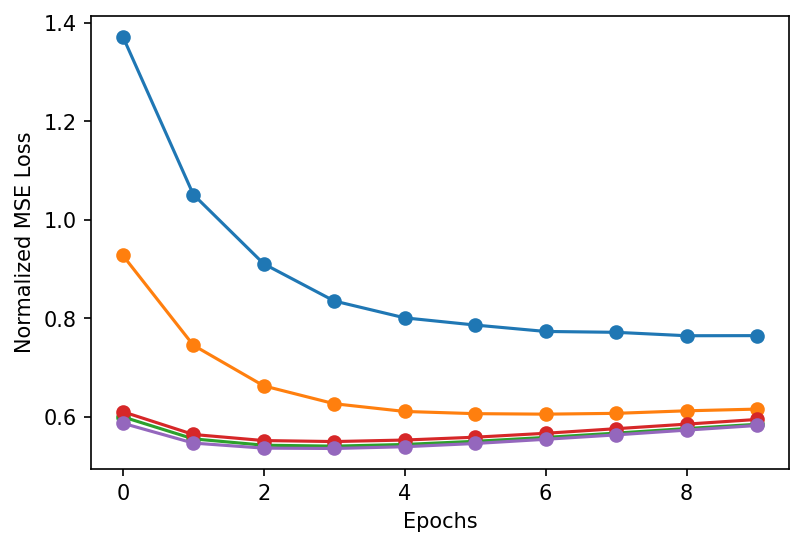}
\caption{Normalized validation learning curves for Linear SynNaMoN on sentence embedding models (blue: BERT, orange: MP-Net, red: GTR-T5-large, green: GTR-T5-base, purple: GTR-T5-xl)}
\label{fig:loss_curves}
\end{figure}

On a less theoretical note, we observe that our learning curves for Linear SynNaMoN on GTR-T5 (Fig. \ref{fig:loss_curves}) are clearly overfitted due to fixing hyperparameters as mentioned in Sec. \ref{sec:experiments}. We remediate this issue in Tab. \ref{tab:val_loss} by reporting the best scores (minimum across epochs) for each learning curve. Since we want to construct the best possible SynNaMoN for a transformer model (as this most accurately reveals the transformer's distillable compositional ability), scores could be slightly improved with further hyperparameter tuning.

\subsubsection{Analysis}

Finally, we explore a single module to determine whether its compositional geometry meets intuitive notions of semantic generalization. Due to methodological difficulties with assessing a single module extracted from our end-to-end training paradigm, we train a Linear module for `NP $\rightarrow$ Det N' on its own. Determiner-noun composition intuitively lies on a spectrum with adjective-noun composition on the other end and quantifier-noun composition in between. While quantifiers like `some' and `all' seem more like determiners, other quantifiers like `several' and `twelve' appear more comparable to adjectives like `swarming' and `grouped'. Intuitively then, we should expect the geometry of quantifier-noun composition to be intermediate to determiners and adjectives.

\begin{figure}[h]
    \centering
    \includegraphics[width=3in]{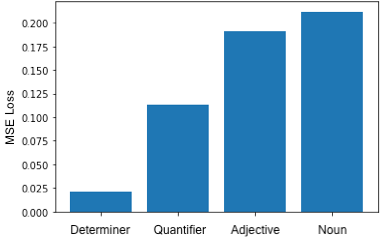}
    \caption{Generalization ability of Determiner Phrase module's linear geometry varies by part-of-speech}
    \label{fig:determiners}
\end{figure}

And this behavior is precisely what we find in our `NP $\rightarrow$ Det N' module. Since it's trained on determiners, it obviously has the lowest MSE for this part-of-speech; we include noun-noun pairs (e.g. `tree cow') as a control. As seen in Fig. \ref{fig:determiners}, the module generalizes to quantifiers intermediately to determiners and adjectives. This demonstrates how SynNaMoN modules may enable interesting analyses of the compositional geometry of syntactic operations in sentence embedding models.

\subsection{Conclusion}
The human ability to apprehend the unitary meaning of a sentence corresponds to a neural model's ability to construct compositional sentence embeddings. In this work, we introduced Syntactic Neural Module Nets and used it in a distillation approach to assess how well syntax explains the sentential semantics computed by a transformer model. We showed that some models are more compositional by this metric, syntax-guided composition is largely linear, and modules learn composition functions that correspond to our semantic intuition. 

Future work could explore this approach's alignment with other compositionality metrics and the non-compositional semantics left uncaptured by SynNaMoNs. We are also interested in how SynNaMoNs of different linguistic formalisms vary in distillability, as well as other potential use cases of SynNaMoNs beyond probing.

\section{Syntax in Attention Flow}
\label{sec:attflow}
In the above work, we saw how distillation allows us to test whether a sentence embedding model's compositional behavior can be described by a syntactic neural module net. However, this does not necessarily prove that the model itself internally implements such syntax-guided compositional structures.

To examine whether this is actually the case, we may visualize the attention flow \cite{abnar2020quantifying} of a sentence embedding model to explore whether it follows patterns predicted by syntax. From an intuitive perspective, the query-key softmax term of self-attention ($\sigma(QK^{\top})V$) modulates the extent to which other words are composed into one word's representation. Thus, from a syntactic perspective, we should expect that words compose within their local clausal boundaries first before composing with distant words. Furthermore, dependencies should contribute more towards the representation of a clausal head than vice versa.

In this work, we do not quantify either of these hypotheses but simply build a visualization tool to explore how attention flow may align with syntax.

\begin{figure}
    \centering
    \subfloat{
    \includegraphics[width=0.48\columnwidth]{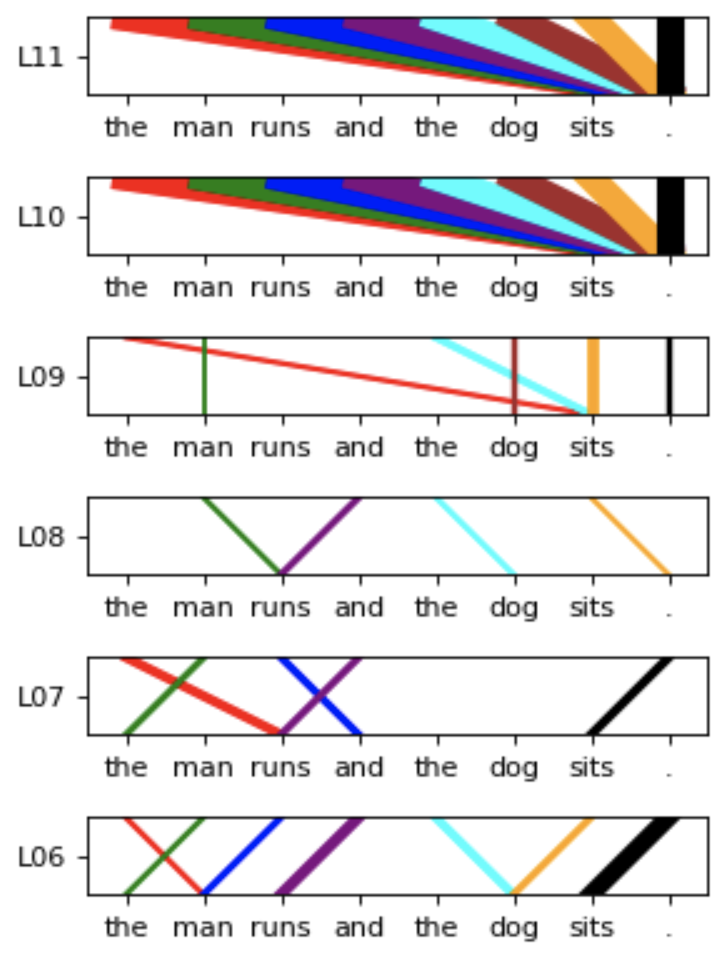}
    \includegraphics[width=0.48\columnwidth]{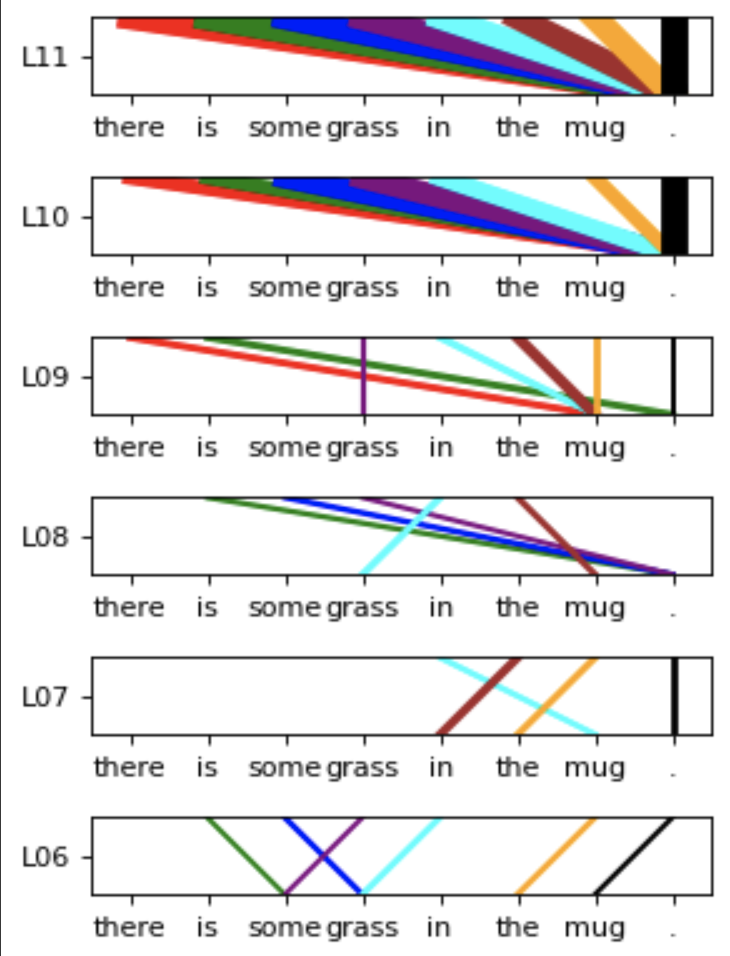}
    }

    \subfloat{
    \includegraphics[width=0.48\columnwidth]{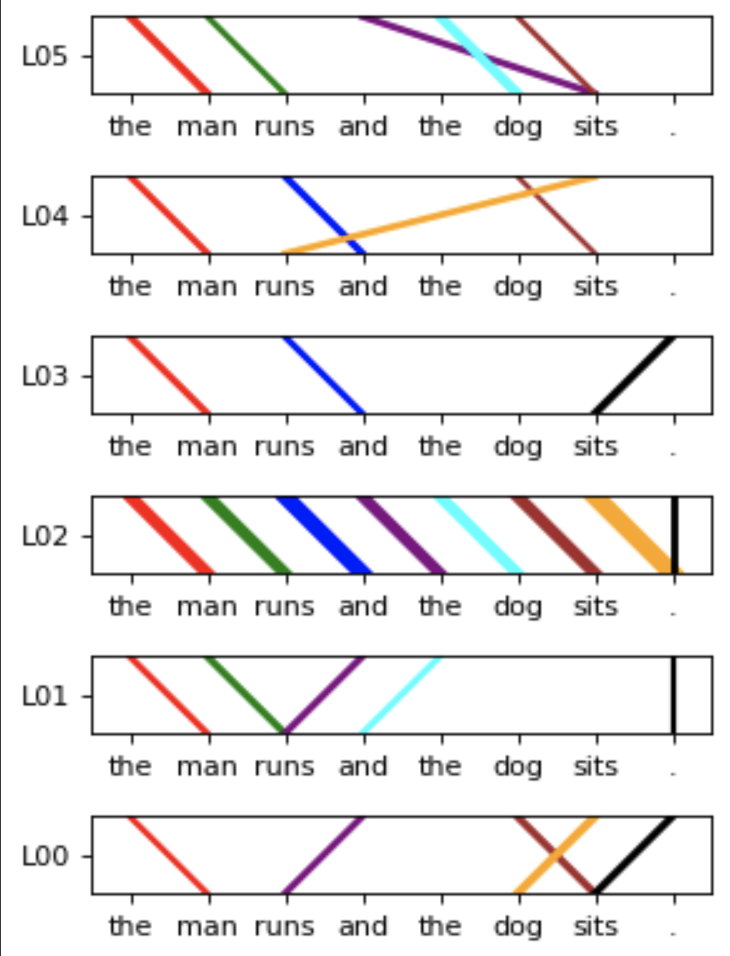}
    \includegraphics[width=0.48\columnwidth]{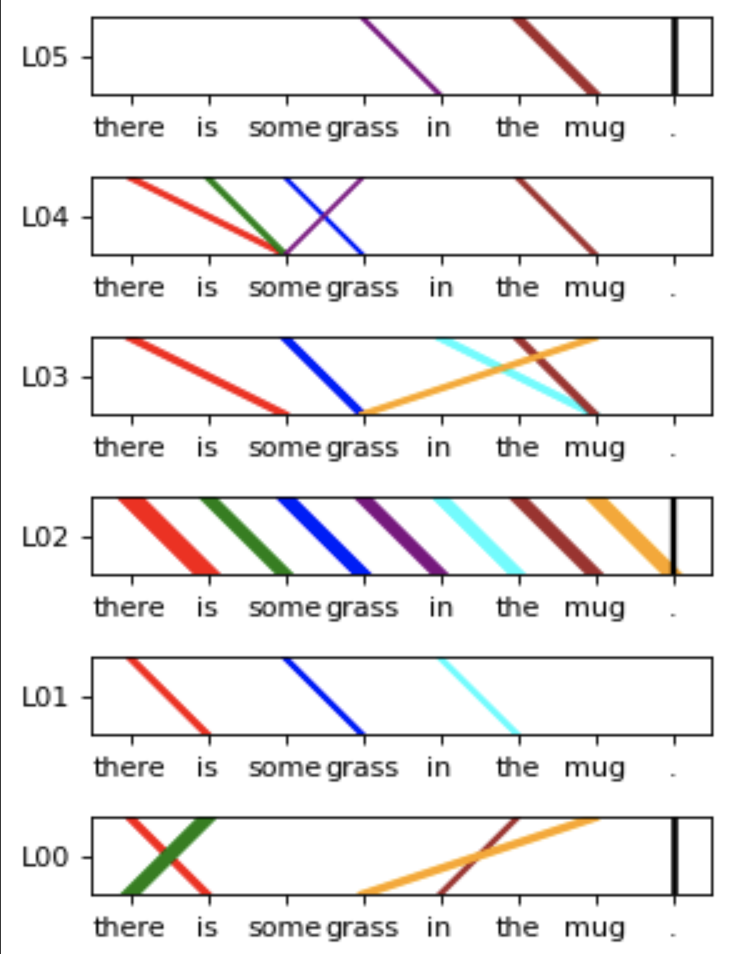}
    }

    \caption{Attention Flow Visualization}
    \label{fig:attflow}
\end{figure}

In Fig. \ref{fig:attflow}, we can observe the attention flow of two sentences through BERT \cite{devlin-etal-2019-bert}. In each column, the self-attention activations averaged across all attention heads (usually visualized as a matrix for each layer) are plotted such that the direction of lines from bottom to the top indicate the words that compose into other words' representations. One may thus follow a word's attention flow from the top down to observe the other words that contribute to its representation and in what order they compose.

The color of the line simply indicates the index of the target word representation while the thickness indicates how strong the attention activation is. Note that not all attention activations are plotted---only those that are greater than $k=2$ standard deviations from the mean attention activation for that layer.

\section[Locating Grounded Composition Circuits with Causal Tracing]{Locating Grounded Composition Circuits with\\Causal Tracing}
\label{sec:vlmechinterp}

Having gained some intuition on how representations are composed in language models, we now turn our attention to rigorously identifying where in a visually grounded language model these compositional representations form. To accomplish this, we adopt a method from causal abstraction analysis \cite{geiger2023causal} and mechanistic interpretability, causal tracing \cite{meng2022locating}.

\begin{figure}[ht]
    \centering
    \includegraphics{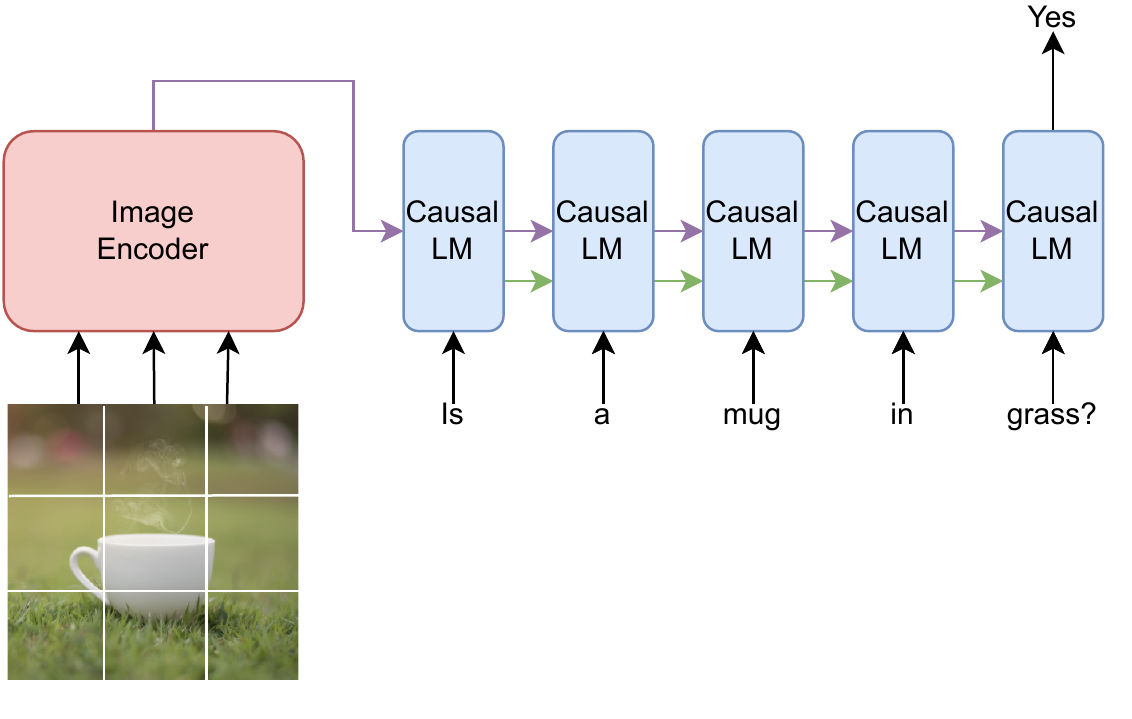}
    \caption{Visual Question Answering using an image-conditioned causal language model. Green arrows are causal attention and purple arrows are cross-attention.}
    \label{fig:image_cap}
\end{figure}

Causal Tracing allows us to identify hidden representations in a neural network that strongly contribute to a specific behavior. In our case, we're interested in the ability of a visually-grounded causal language model to answer questions that depend on the compositional structure of an image. 

In Causal Tracing, we perform a normal forward pass as shown in Fig. \ref{fig:image_cap}, as well as a forward pass where we corrupt part of the input---in our case, the image. Then, we patch each hidden representation from the normal forward pass into the corrupted forward pass to see which hidden representations we patch result in the most significant improvement in performance (answering yes or no). To verify that we're identifying neural states relevant for compositional representation and not just for answering ``yes'' or ``no'', we can perform a similar operation on Winoground pairs.

\chapter{Enabling models to compose visually grounded semantics}

\section{Past Work}
Compositionality is the ability to combine meanings of constituents according to structured rules. Recent work shows that Vision-Language Models (VLMs) fail to construct compositional representations and generally ignore syntactic \& structural information \cite{thrush22, milewski22, liang2022foundations}.
Winoground \cite{thrush22} is a vision-language compositionality task that tests a VLM's ability to match syntactic permutations of text with their visual interpretations, for example correctly matching ``grass in mug'' and ``mug in grass'' to their corresponding images. Winoground finds that all recent state-of-the-art VLMs perform below chance levels on this compositionality task. Contemporaneously, \citet{milewski22} probe for structural knowledge in VLMs, finding that they encode significantly less linguistic syntax than Language Models (LMs) and virtually no visual structure. Recently, \citet{yuksekgonul2022bags} built a large dataset confirming that VLMs treat images as a `bag of objects' and don't adequately represent visuo-linguistic relations.

Since models must determine whether the compositional structure of an image matches that of the caption, it's important for the model to learn to cross-modally align intra-modal relations. That is, if the relation from `mug' to `grass' is `in-ness', the model should recognize when the equivalent physical relation holds between a mug and grass in the image, and representationally align these relations such that an image-text matching head may more easily determine whether the relations are cross-modally equivalent. In simpler terms, the compositional structure of input for each modality should be represented such that they can be cross-modally matched even for difficult examples like Winoground.

\begin{figure}[h]
    \centering
    \includegraphics[width=0.5\columnwidth]{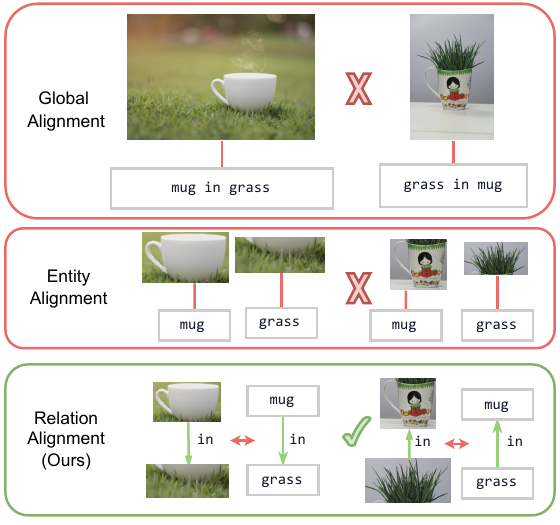}
    \caption{Global Alignment (GA) only aligns the entire image with the corresponding caption. Entity Alignment (EA) extracts entities from the image and caption for finer-grained alignment. Relation Alignment (RA) cross-modally aligns the intra-modal relations between entities in both the image and the text. We show RA is vital to improve compositional performance.}
    \label{fig:wino_overview}
\end{figure}

Unfortunately, there has been less highly influential work on \textbf{relation alignment} between vision \& language, and \citet{thrush22} did not benchmark any such models. In this work, we begin exploration of these relation alignment approaches by tentatively grouping them into 3 categories: 
\begin{enumerate}
    \item Structural Data: training a model on data that explicitly captures relational structure \cite{Wu_2019_univse, zhang2019agha, yu20ernievil, cui21rosita, wan2021cliora, khan2022simla}
    \item Structural Model: infusing an inductive bias into the architecture of the model that enables more compositional representations \cite{andreas2016neural, guo19vsua, hong2021vlgrammar, zhang-2022-improve, wang2022sgeitl, kim2022cross, wang22vqagnn}
    \item Structural Training: modifying the objective function or imposing a parameter constraint to encourage relation alignment \cite{ren21iais, yang21apn, yang21catt, xue2021imf}
\end{enumerate}

While some of these works introduce ideas from multiple of these categories, we group them by their core contribution. For example, ROSITA proposes a graphical data pre-training approach, and a self-supervised objective to accompany it; we consider it a Structural Data approach since the training objective ultimately is just a necessity for the data being provided.

Unfortunately, many of these works do not provide publicly available code or pre-trained checkpoints, so we were unable to complete an exhaustive analysis of the compositional performance of these relation alignment approaches. Due to the added complexity of Structural Model approaches, we leave exploration of their compositional abilities to future work.

Since Structural Data approaches require complex annotations and Structural Model approaches are often incompatible with large transformers, we identify Structural Training as a promising avenue for providing compositional inductive biases to VLMs due to their architecture-agnostic compatibility and computational scalability.

\section{Cross-modal Attention Congruence Regularization}
\label{sec:cacr}

We chose one exemplar for both Structural Data (ROSITA) and Structural Training (IAIS) that made their pre-trained image-text matching checkpoints available; we generated their scores on Winoground, which have not previously been calculated. In Tab. \ref{table:wg_alignment}, we present these two relation alignment models' Winoground scores alongside a few entity alignment and global alignment models.

\begin{table}[h]
\centering
\begin{tabular}{ p{3.4cm}||p{1cm} p{1cm} p{1cm} }
 \hline
 Model &Text &Image &Group\\
 \hline
 MTurk Human & 89.50 & 88.50 & 85.50\\
 \hline
 IAIS (RA-ST) & $\mathbf{42.50}$ & $\mathbf{19.75}$ & $\mathbf{16.00}$\\
 OSCAR+ (EA) & 37.75 & 17.75 & 14.50\\
 ROSITA (RA-SD) & 35.25 & 15.25 & 12.25\\
 UNITER (EA) & 38.00 & 14.00 & 10.50\\
 CLIP (GA) & 30.75 & 10.50 & 8.00\\
 LXMERT (GA) & 19.25 & 7.00 & 4.00\\
 \hline
\end{tabular}
\caption{Comparison of Winoground scores for models using Global Alignment (GA), Entity Alignment (EA), Relation Alignment with Structural Data (RA-SD), and Relation Alignment with Structural Training (RA-ST). We find that IAIS, a recent relation alignment approach that uses attention regularization for structural training achieves universal performance improvements.}
\label{table:wg_alignment}
\end{table}

Notice that global alignment approaches tend to perform the lowest on Winoground, even when scaled considerably. Entity alignment approaches perform intermediately and OSCAR+ specifically held the state-of-the-art prior to our benchmarking of these relation alignment models. Of the two relation alignment approaches we benchmark, IAIS beats out OSCAR+ and achieves a new state-of-the-art on Winoground. But ROSITA, despite providing structural data to encourage cross-modal relation alignment, underperforms OSCAR+. We attribute this partly to the improved visual features OSCAR+ has access to as a result of VinVL, but further comparison of IAIS and ROSITA is explored in our recent work.

In this work, we propose a Structural Training approach for relation alignment that uses the cross-modal attention matrix as a change of basis\footnote{not defined in a strict linear algebraic sense} to the opposite modality, which we then compare to the original modality to calculate a divergence loss, effectively measuring cross-modal congruence between intra-modal attentions. 

We show how our approach, Cross-modal Attention Congruence Regularization (CACR), generalizes previous Structural Training work on cross-modal attention regularization (IAIS \cite{ren21iais}) by taking into account all possible entity alignments and computationally simplifying relation alignment. The CACR regularization term can easily be dropped into most transformer-based Vision-Language model objectives with no added data and minimal computational overhead, to encourage relation alignment during training. Finally, we show that CACR$_{\text{base}}$ improves on IAIS$_{\text{base}}$—where IAIS$_{\text{large}}$ holds the current state-of-the-art on Winoground.

\subsection{Method}

How can we infuse the vision-language model's training objective with an implicit structural prior that encourages cross-modal alignment of relations?

To attempt a solution to this question, we begin by noting that attention activations encode some degree of relational information. Attention values in transformers may be seen as an informational gating mechanism that implicitly encode how representations are composed \cite{abnar2020quantifying}. For example, past work in language has shown how syntax trees may be extracted \cite{marevcek2019balustrades} from attention across layers and used to guide attention \cite{bai2021syntax, li2020improving} for improved compositionality. In this section, we extend this intuition to the multimodal domain by proposing to use the cross-modal attentions, which as a change-of-basis matrix encode a transformation from one modality's compositional structure to the opposite modality's, to encourage cross-modal relation alignment.

\subsubsection{Relation Alignment Using Attention}
In specific, we focus on the self-attention matrix $S$ computed in a transformer by 
\begin{equation}
    S = QK^\top = (X W^Q)(X W^K)^\top
\end{equation}

Then, some row $i$ in S corresponds to a distribution over columns $j_0, ..., j_n$ where $S_{i,j}$ tells us how much of the previous layer's entity representation $j$ we want to infuse into the current layer's entity representation $i$, intuitively their compositional relation. Since $X$ is a series of visual and linguistic tokens, we can segment $S$ into four submatrices for intra- and cross-modal relations \cite{bugliarello-etal-2021-multimodal}. Denote the intra-modal attention submatrices in the last multimodal encoder layer as $S_{VV}$ (vision to vision) and $S_{LL}$ (language to language); the cross-modal attention matrices as $S_{VL}$ (vision to language) and $S_{LV}$ (language to vision).

\begin{equation}
    S =
\begin{pmatrix}
S_{LL} & S_{LV} \\
S_{VL} & S_{VV}
\end{pmatrix}
\end{equation}

If an image and caption have the same underlying compositional structure, the entities that cross-modally correspond to each other should bear similar intra-modal compositional structure. That is, a word $w$ should attend to other words (in $S_{LL}$) in a similar way that its visual object counterpart $o$ attends to other objects (in $S_{VV}$). Furthermore, we can use the cross-modal matrices ($S_{LV}$ and $S_{VL}$) to identify entities that cross-modally correspond as they will generally attend to each other \cite{aflalo2022interpret}. Unfortunately, since representations are heavily contextualized by the final layer, clear bijective correspondences between words and objects may not always be identified using an argmax over the cross-modal attention matrix as \citet{ren21iais} attempts. Deeper analysis of when their model, IAIS, fails to identify cross-modal bijective correspondences is provided in Sec. \ref{sec:analysis}.

\subsubsection{Attention Congruence}

\begin{figure*}
    \centering
    \includegraphics[width=\linewidth]{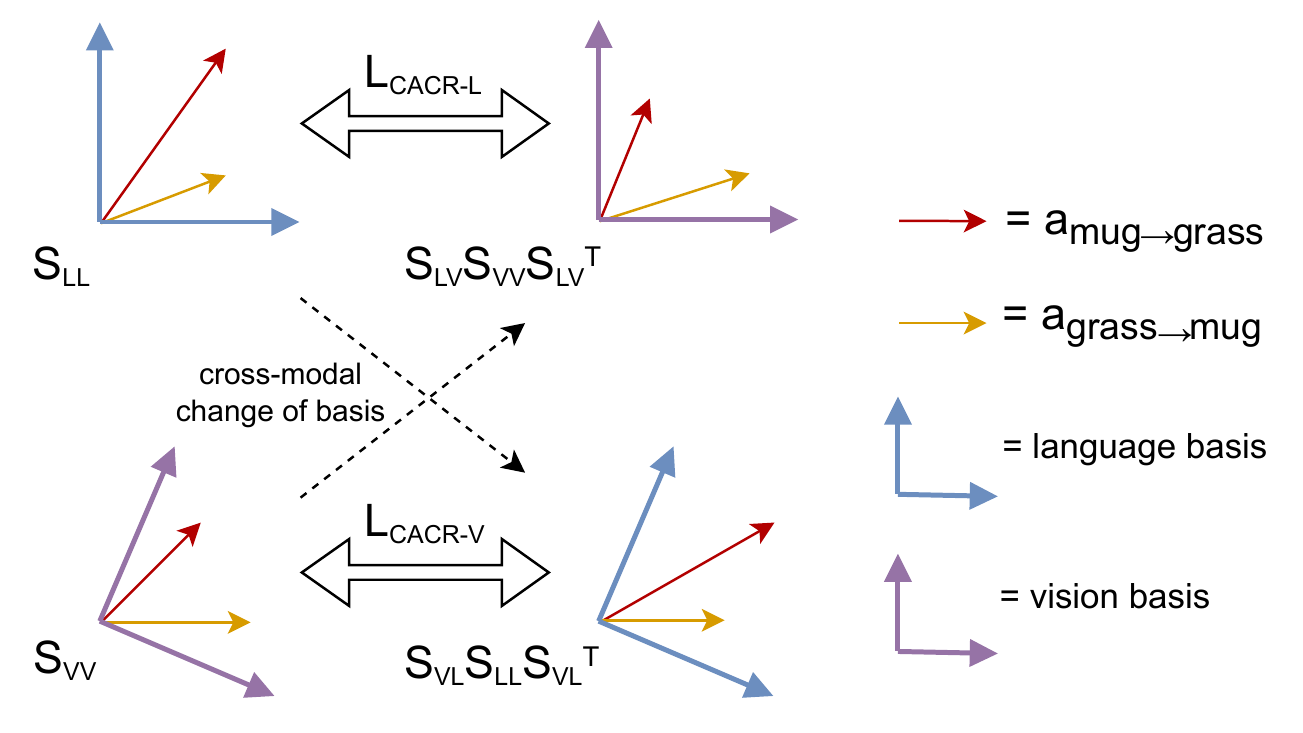}
    \caption{Top: language attention ($S_{LL}$) is aligned with the visual attention projected into the language basis ($S_{LV}S_{VV}S_{LV}^\top$) to calculate $\mathcal{L}_{CACR-L}$; specific attention values (yellow, red) capturing intra-modal relations are cross-modally aligned as a result. Bottom: as above, but in the vision basis.}
    \label{fig:basis}
\end{figure*}

We opt to use the cross-modal matrices ($S_{LV}$ and $S_{VL}$) as a whole to `change basis' to the opposite modality, with which we can then calculate `congruence' with the original modality. However, we use `change of basis' and `congruence' loosely since the cross-modal matrices are not guaranteed to be square and thus do not satisfy strict linear algebraic definitions. We formulate $S_{VV}$ in the language basis as $S_{LV}S_{VV}S_{LV}^\top$, which we then encourage to be similar to $S_{LL}$.

Under the hood, this says that for each $a_{i \rightarrow j} \in S_{LL}$, we can use row vectors $S_{LV,i}$ and $S_{LV,j}^\top$ to calculate a weighted sum $a_{i \rightarrow j}^*$ over $S_{VV}$. If we were to do this for all $i,j$, we would construct a matrix of the same shape as $S_{LL}$ where each entry is $a_{i \rightarrow j}^*$, i.e. an approximation of the visual correspondent of the relation $a_{i \rightarrow j}$ taking into account all the possible cross-modal alignments of $i$ and $j$. Since this computation intuitively makes a lot of sense and may more easily be compared to previous approaches, we choose to illustrate it in Fig. \ref{fig:attention}. However, since this computation is relatively expensive, we instead use the $S_{LV}S_{VV}S_{LV}^\top$ formulation which produces the same matrix of $a_{i \rightarrow j}^*$ values but with considerably fewer operations. This also enables us to view the operation as a `change-of-basis' to the opposite modality and the CACR loss as encouraging a sense of cross-modal `congruence'.

Specifically, we align the original $S_{LL}$ with the language-basis $S_{VV}$ matrix using $\mathcal{L}_{\text{CACR-L}}$:
\begin{equation}
    \mathcal{L}_{\text{CACR-L}} = \text{m-KL}(\sigma(S_{LV}S_{VV}S_{LV}^\top), \sigma(S_{LL})).
\end{equation}

We apply a softmax to normalize both matrices since $S_{LV}S_{VV}S_{LV}^\top$ will generally be larger in scale due to summation. Additionally, m-KL$(\cdot)$~\cite{ren21iais} is a symmetric matrix-based Kullback-Leibler Divergence (m-KL) which measures the distance between two matrices $S$ and $S'$:
\begin{equation}
    \text{m-KL}(S, S') = \sum_i^N \text{KL}(S_i||S_i') + \text{KL}(S_i'||S_i),
\end{equation}
where $(\cdot)_i$ stands for the $i^{\text{th}}$ row-vector in the matrix.

Similarly, we have $\mathcal{L}_{\text{CACR-V}}$:
\begin{equation}
    \mathcal{L}_{\text{CACR-V}} = \text{m-KL}(\sigma(S_{VL}S_{LL}S_{VL}^\top), \sigma(S_{VV})),
\end{equation}

Combining $\mathcal{L}_{\text{CACR-V}}$ and $\mathcal{L}_{\text{CACR-L}}$, we present our $\mathcal{L}_{\text{CACR}}$ objective, an attention activation regularizer for cross-modal relation alignment:
\begin{equation}\label{eq:loss}
    \mathcal{L}_{\text{CACR}} = \mathcal{L}_{\text{CACR-V}} + \mathcal{L}_{\text{CACR-L}}.
\end{equation}

When the vision inputs and the language inputs have the same sequence length and $S_{VL}, S_{LV}$ are invertible, then $S_{VV}$ and $S_{VL}S_{LL}S_{VL}^\top$ (as well as $S_{LL}$ and $S_{LV}S_{VV}S_{LV}^\top$) can become strictly congruent. In this case, $S_{VL}S_{LL}S_{VL}^\top$ can be interpreted as the language view of $S_{VV}$. Aligning $S_{VL}S_{LL}S_{VL}^\top$ and $S_{VV}$ leads to cross-modal relation alignment. It is similar for $S_{LV}S_{VV}S_{LV}^\top$ and $S_{LL}$. In the general case where the vision inputs and the language inputs may have different sequence lengths, the two forms are not linear algebraically congruent but the relevant intuition still holds.

\subsubsection{Hard and Soft Cross-modal Equivalence}

\begin{figure*}
    \centering
    \includegraphics[width=\linewidth]{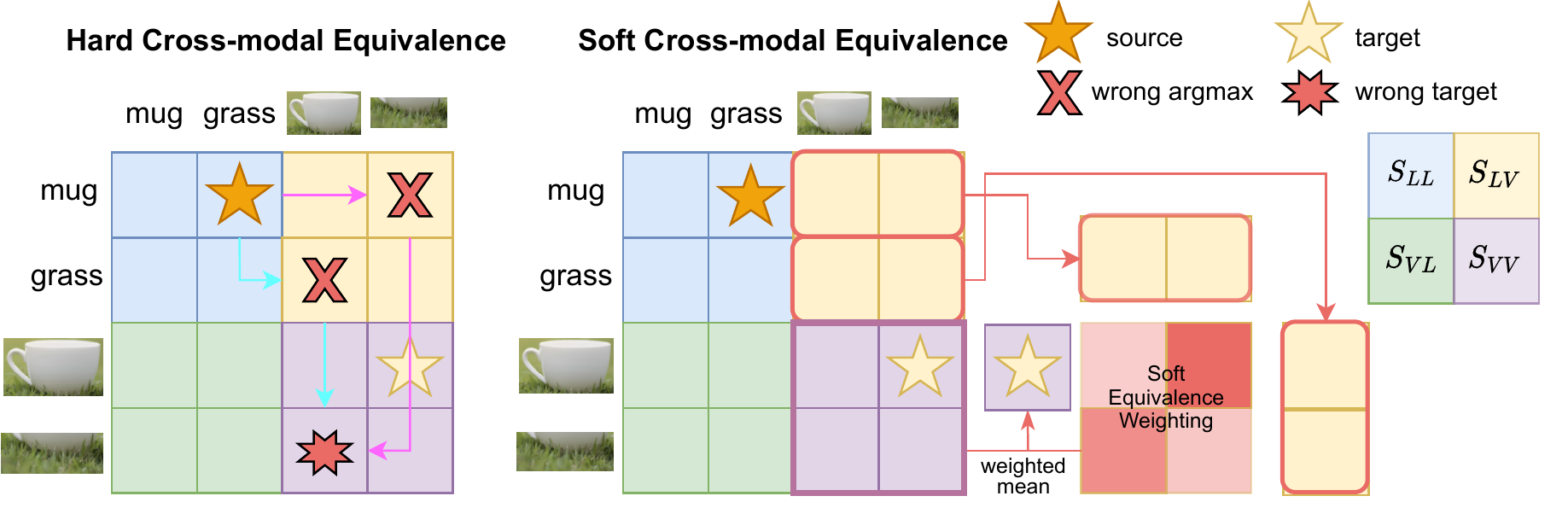}
    \caption{Comparison of the hard cross-modal equivalence used in IAIS (left) and the soft cross-modal equivalence we propose in CACR (right), with an example from Winoground to illustrate how the target visual equivalent (yellow star) of a source linguistic relation (orange star) is calculated. Hard cross-modal equivalence attempts to build a one-to-one mapping between language and vision by applying argmax (red crosses) to the $S_{LV}$ row vectors. Our soft cross-modal equivalence instead uses the whole $S_{LV}$ row vectors to calculate a weighted (red) mean over $S_{VV}$. The scalar that is produced corresponds to the attention from `mug' to `grass' but in a visual basis. We note that IAIS can be seen as a special case of soft cross-modal equivalence by forcing the attention matrix (red) to be a one-hot matrix where the max value is set to 1 and all others to 0. CACR linear algebraically simplifies soft cross-modal equivalence for computational efficiency.}
    \label{fig:attention}
\end{figure*}

In this section, we show that CACR can be interpreted as leveraging cross-modal soft equivalences, where IAIS \cite{ren21iais} uses hard bijective equivalences. In their approach, each element in the intra-modal attention matrix is aligned with a single counterpart in the opposite modality. This is built upon a strict assumption that there exists a one-to-one mapping (provided by an argmax over the cross-modal attention) from $S_{LL}$ to $S_{VV}$ and vice versa, which is unsatisfied in practical cases. CACR may be seen as a soft cross-modal equivalence method which instead uses the whole $S_{LV}$ (or $S_{VL}$) to implicitly build an `equivalence weighting' which is then used to compute a weighted mean over $S_{VV}$ (or $S_{LL}$). We illustrate and compare hard cross-modal equivalence and our soft cross-modal equivalence in Figure~\ref{fig:attention}, taking the language-side alignment as an example. 

We note that IAIS could be seen as a special case of soft cross-modal equivalence by forcing the cross-modal attention map to be a one-hot matrix, i.e., taking the argmax of the attention matrix as the index of the cross-modal counterpart. We show in Section~\ref{sec:analysis} that IAIS can have inferior performance when a clear bijective cross-modal correspondence isn't available.

In Alg.~\ref{alg:siaisl}, we show the pseudo-code of the soft cross-modal equivalence method for calculating the vision-side loss. $\mathcal{S}_{CACR-V}$ can be computed similarly. Computing the hard and soft cross-modal equivalence is computationally complex and difficult to be parallelized due to indexing operations. For practical applications, we sought to simplify this soft cross-modal equivalence algorithm to a mathematical equivalent that would improve computational tractability. From here, we arrive at CACR, which is a closed-form formulation of soft cross-modal equivalence which utilizes only differentiable matrix multiplications. Therefore, our CACR is more \textbf{computationally efficient} and \textbf{easier to be parallelized} than soft cross-modal equivalence. 

\begin{algorithm}
\setstretch{1.2}
\caption{Soft Cross-modal Equivalence (V)}
\label{alg:siaisl}

\begin{algorithmic}[1]
\Require $S_{LL} \in N \times N, S_{VL} \in N \times M, S_{VV} \in M \times M$
\State $\mathcal{L} \gets 0$
\For {$i, j \in S_{VV}$}
\State $W \gets S_{VL}[i] \cdot S_{VL}^\top[j]$ \Comment{soft weighting}
\State $a_{i \rightarrow j}^* \gets \overline{W \circ S_{LL}}$ \Comment{element-wise weighted mean}
\State $\mathcal{L} = \mathcal{L} + \text{m-KL}(a_{i \rightarrow j}^*, S_{VV}[i,j])$
\EndFor
\State \Return $\mathcal{L}$
\end{algorithmic}
\end{algorithm}

\subsubsection{Proof of Equivalence Between CACR and Soft Cross-modal Equivalence}
Computing the hard (IAIS) and soft cross-modal equivalence is computationally complex and difficult to parallelize due to indexing operations. However, CACR loss is mathematically equivalent to soft cross-modal equivalence but can be computed efficiently. We take $\text{CACR}_{V}$ for illustration of this equivalence, but $\text{CACR}_{L}$ can be proved in the same way.

Beginning with the visual-basis form of $S_{LL}$ in CACR, the attention at index $[i, j]$ in $S_{VL}S_{LL}S_{VL}^\top$ is
\begin{equation}\label{eq:proof}
    \begin{array}{l}
        (S_{VL}S_{LL}S_{VL}^\top)[i,j] \\[4pt]
        
        = \sum_p^{N_L} \sum_k^{N_L} a_{v_i\rightarrow l_k} a_{l_p \rightarrow l_k} a_{v_j \rightarrow l_p}  \\[4pt]
        
        = \sum_p^{N_L} \sum_k^{N_L} \underbrace{S_{VL}[i, k] S_{VL}[j, p]}_{\text{soft weighting}} S_{LL}[p, k]
    \end{array}
\end{equation}

        
        

where $a_{v_i\rightarrow l_j}$ stands for the attention from the $i$-th visual token to the $j$-th linguistic token, $N_L$ is the total number of language tokens and $N_V$ is the total number of the visual tokens. Comparing Eq.\ref{eq:proof} and Alg.\ref{alg:siaisl}, we observe that the summation we arrive at above is equivalent to the content of the for-loop (line 3-5). Thus, although of seemingly different linear algebraic form, CACR generalizes IAIS by way of its equivalence to the Soft Cross-modal Equivalence formulation presented above.

\subsection{Results}

How does CACR compare to other vision-language models in its compositional ability?

\begin{table}[h]
\centering
\begin{tabular}{ p{2.8cm}||p{1cm} p{1cm} p{1cm} }
 \hline
 Model &Text &Image &Group\\
 \hline
 MTurk Human & 89.50 & 88.50 & 85.50\\
 \hline
 CACR$_{\text{base}}$ & 39.25 & \textbf{17.75} & \textbf{14.25}\\
 UNITER$_{\text{large}}$ & \textbf{43.50} & 14.75 & 13.75\\
 IAIS$_{\text{base}}$ & $37.50$ & $16.75$ & 13.00\\
 UNITER$_{\text{base}}$ & 32.75 & 11.75 & 8.50 \\
 \hline
\end{tabular}
\caption{CACR outperforms its pre-trained baseline (UNITER) and an alternative attention regularization approach (IAIS) across all Winoground scores.}
\label{table:wg_scores}
\end{table}

In Tab. \ref{table:wg_scores}, we present our approach's scores alongside a few other models. Since we use CACR to fine-tune UNITER, we include scores for the two baseline UNITER sizes. We also include scores for $\text{IAIS}_{\text{base}}$ which is also built on UNITER. 

The fact that $\text{CACR}_{\text{base}}$ outperforms $\text{IAIS}_{\text{base}}$ suggests that, with adequate computational resources, $\text{CACR}_{\text{large}}$ could similarly outperform $\text{IAIS}_{\text{large}}$, potentially achieving a new state-of-the-art on Winoground. Furthermore, its performance compared to $\text{UNITER}_{\text{large}}$ is impressive considering that $\text{CACR}_{\text{base}}$ is approximately half its size in parameters.

\begin{table}[h]
\centering
\begin{tabular}{ p{2.2cm}||p{0.9cm} p{0.9cm} p{0.9cm} p{0.9cm}}
 \hline
 Model &Image R@1 &Image R@10 &Text R@1 &Text R@10\\
 \hline
 IAIS$_{\text{base}}$ & 73.54 & 96.32 & 86.10 & 99.10 \\
 UNITER$_{\text{base}}$ & 72.52 & 96.08 & 85.90 & 98.80 \\
 CACR$_{\text{base}}$ & 70.88 & 95.68 & 83.50 & 98.80 \\
 \hline
\end{tabular}
\caption{CACR performance on Flickr30k has marginal reductions from UNITER, suggesting performance could be improved even further with a hyperparameter search.}
\label{table:flickr_scores}
\end{table}

Finally, we report Flickr30k retrieval scores in Tab. \ref{table:flickr_scores} to verify that we are not somehow overfitting to Winoground. Though CACR takes some minor losses to its retrieval scores, this may be attributed to imperfect hyperparameters, suggesting that CACR's performance on Winoground could be even higher with adequate hyperparameter tuning. It's also important to remember here that we're only training on Flickr30k, so this isn't a case of our model overfitting to Winoground and `forgetting' its true image-text matching ability. Rather, it shows that the hyperparameters that we adapted from IAIS need to be modified to more perfectly train CACR on Flickr30k, which would then carry over to compositional improvements on Winoground.

We fine-tuned CACR on Flickr30k \cite{young2014flickr} for 5000 epochs using PyTorch \cite{paszke2019pytorch} with a train-validation-test split of 90-5-5. The training batch size is 4 and 31 negative samples are provided for every individual positive sample in a standard image-text matching training setup. We use a learning rate of $5 \times 10^{-5}$, the $AdamW$ optimizer \cite{loshchilov2017decoupled}, and introduce $\mathcal{L}_{\text{CACR}}$ with an exponential warmup schedule. Training was completed on a node with 4 NVIDIA GTX 1080 Ti's, each with 11 GB of memory.

\subsection{Analysis}
\label{sec:analysis}
Why does CACR's soft cross-modal equivalence approach outperform hard cross-modal equivalence?




\subsubsection{Qualitative}

Hard cross-modal equivalence, implemented by IAIS, assumes that cross-modal submatrices can be used to find a singular equivalent of an entity in the opposite modality. Specifically, if $i^* = \text{argmax}(S_{LV}[i])$ then $S_{LL}[i]$ should correspond to $S_{VV}[i^*]$. In simple terms, IAIS says the following: if word A attends most to object A and word B attends most to object B, then word A should attend to word B in a similar way that object A attends to object B. Underlying IAIS is the hard assumption that argmaxing over the cross-modal attention submatrix is an effective means of identifying the opposite modality equivalent of an entity. However, we show in this section that this is often not the case. 

\begin{figure}
    \centering
    \subfloat[\centering \textit{clouds}]{{\includegraphics[width=0.49\columnwidth]{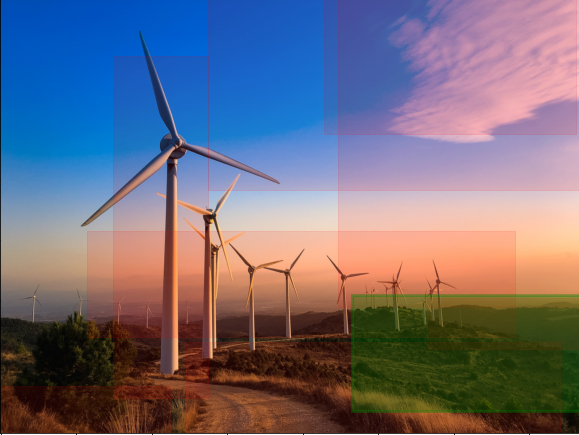} }}%
    \subfloat[\centering \textit{turbines}]{{\includegraphics[width=0.49\columnwidth]{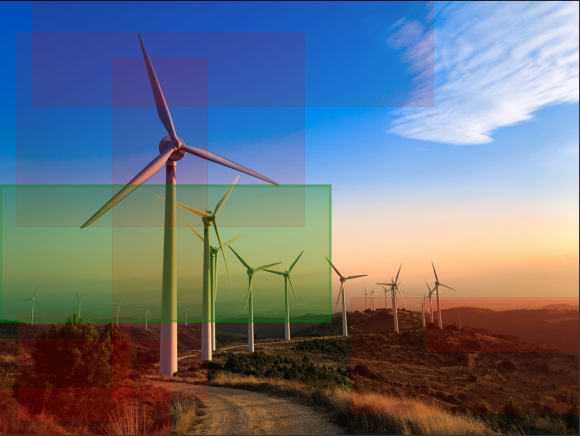} }}%
    
    \subfloat[\centering \textit{``hammering something together"}]{{\includegraphics[width=0.49\columnwidth]{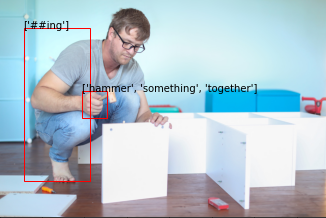} }}
    \subfloat[\centering \textit{``together hammering something"}]{{\includegraphics[width=0.49\columnwidth]{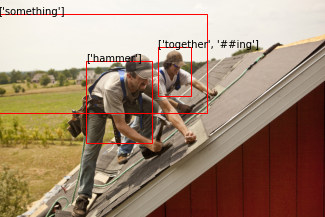} }}

    \caption{Top: UNITER $S_{LV}$ attention for caption \textit{``a few clouds and many wind turbines"}, with the bounding box maximally attended to by the token in green; other highly attended boxes in red. Bottom: UNITER $S_{LV}$ attention with bounding boxes labeled with the tokens that maximally attend to them. Note that argmaxes often fail to precisely identify cross-modal equivalence.}
    \label{fig:slv_ex}
\end{figure}

Given the argmaxes for rows in the $S_{LV}$ submatrix, we can identify the bounding box that each token maximally attends to, which IAIS assumes is its visual equivalent. In Fig. \ref{fig:slv_ex}a, we visualize an example where `clouds' maximally attends (green) to the ground, which would prevent IAIS from identifying the correct cross-modal equivalence. `Turbines' (Fig. \ref{fig:slv_ex}b), on the other hand, maximally attends to a bounding box that better matches our intuition. It is qualitatively clear from the several examples displayed that the argmax assumption often fails to identify the correct cross-modal equivalence. Since words may attend to several visual tokens for different reasons, we shouldn't assume that the cross-modal argmax provides us with a clear bijective correspondence.


Instead, the cross-modal matrices should be seen as providing useful high-level information about what visual entities are relevant to a word, and vice versa. We can certainly gain useful information about cross-modal correspondences using it, but it isn't as simple as using an argmax, due to words having multiple referents and entity representations being intermixed. Instead, our soft cross-modal equivalence approach takes all the possible cross-modal alignments into account with a weighted sum.

\begin{figure}[ht]
    \centering
    \subfloat[\centering \textit{dog}]{{\includegraphics[width=0.49\columnwidth]{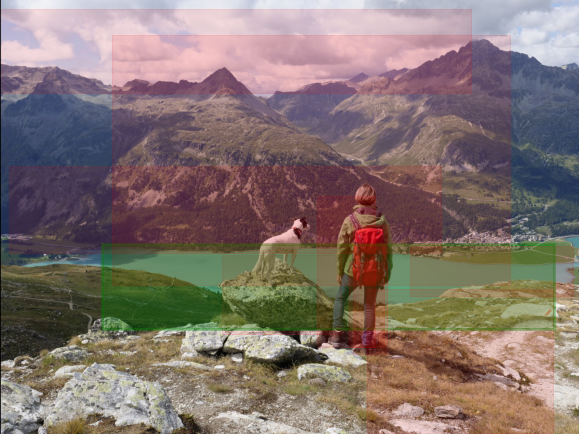} }}%
    \subfloat[\centering \textit{person}]{{\includegraphics[width=0.49\columnwidth]{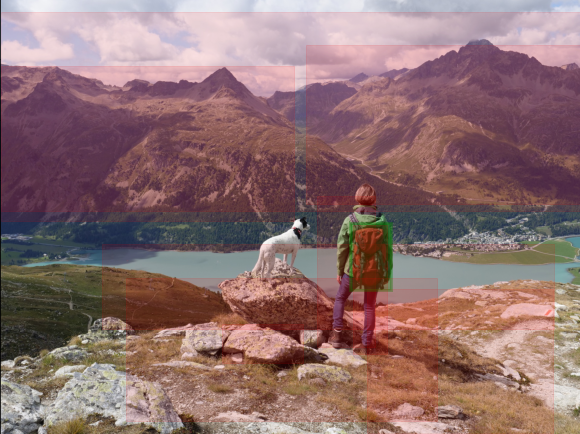} }}%

    \subfloat[\centering  \textit{mug}]{{\includegraphics[width=0.49\columnwidth]{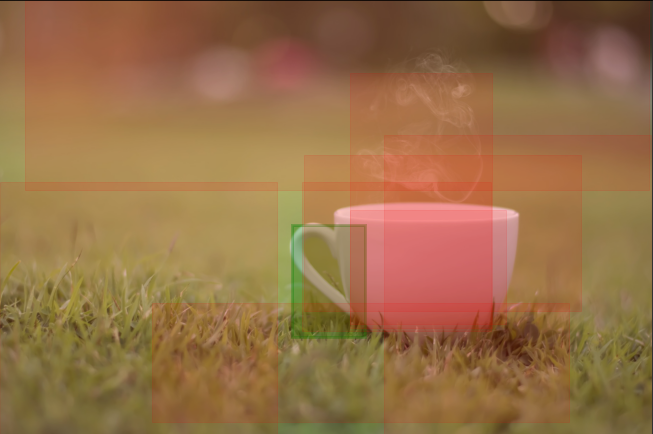} }}
    \subfloat[\centering  \textit{grass}]{{\includegraphics[width=0.49\columnwidth]{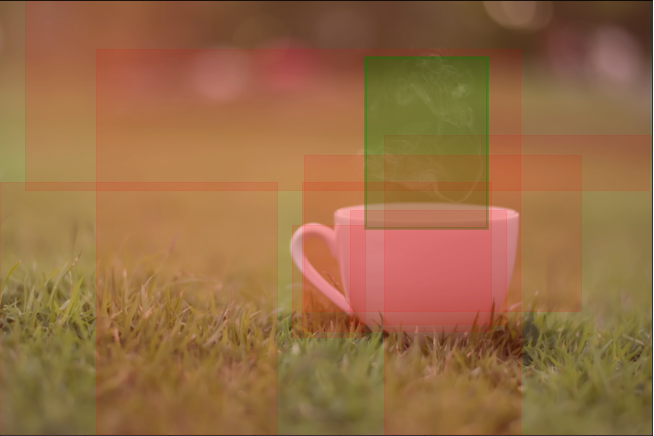} }}
    
    \caption{UNITER $S_{VL}$ attention for captions \textit{``a dog on a rock next to a person"} and \textit{``there is a mug in some grass"}. Shown are boxes that attend highly to the displayed token, with the maximally attending bounding box in green; others in red. Observe that although the argmax often does pick up on a relevant bounding box, it is prone to missing critical visual information, e.g. focusing on only the backpack in (b).}
    \label{fig:svl_ex}
\end{figure}

To illustrate how the soft approach accounts for critical cross-modal alignment information, we present a few Winoground examples with UNITER's cross-modal attention activations in Fig. \ref{fig:slv_ex} and \ref{fig:svl_ex}. We use UNITER since this is the baseline model from which attentional information is bootstrapped to calculate cross-modal alignments. For example, in Fig. \ref{fig:svl_ex}c, using the representation for the bounding box covering the mug's handle may not adequately capture the visual referent of `mug' and therefore disrupt our ability to calculate the visual-basis relation between `mug' and `grass' if restricted by an argmax.

\subsubsection{Quantitative}
In the absence of annotations, we attempted a quantitative measurement of whether overlap in argmaxes (several words attending to one bounding box or vice versa) as quantified by the Shannon Entropy of argmax indices inversely correlates with soft Winoground score. Intuitively, if an example has more like a one-to-one mapping between text and image, the entropy of its cross-modal argmaxes should be higher as each token will attend to a different box, which would suggest that the model is better aligning entities. However, we found no significant correlation with Winoground score, which we attribute to the fact that high entropy on its own doesn't mean \textit{correct} entity alignment. 

Rather, high entropy in argmax indices could still be produced by a bad representation if `mug' attends to the grass \& `grass' attends to the mug; conversely, low entropy could be produced by a good representation for an example like `fire truck' where two tokens refer to a single object. Quantitative exploration of cross-modal attention is difficult without annotations and we leave this task to future work to explore in a multimodal compositionality context.

As a general takeaway, while the cross-modal argmax assumption of IAIS does hold in some cases and may be more meaningful during the course of IAIS training, it is clearly quite a strict assumption that could suffer if an entity attends to several cross-modal entities or there are no corresponding cross-modal entities. Furthermore, since IAIS is only active in the final self-attention layer, all the token representations are intermixed and therefore don't necessarily have a one-to-one correspondence with our intuitive notions of what they should be—the word `turbine' may not solely represent the traditional meaning of that word but perhaps the entire scene that includes the turbines, clouds, and ground.

We hypothesize that by removing the hard argmax assumption, our approach better accounts for varying cross-modal entity equivalences and thus enables stronger relation alignment. By also calculating alignment between all pairs of source and target modality entities, CACR should considerably improve sample efficiency, which is important considering that the final layer $S$ matrix of the converged IAIS model is largely flat. Therefore it's important to backpropagate as much alignment knowledge over the course of training as possible, which CACR's soft equivalence weighting implicitly enables.

\subsection{Conclusion}

In this work, we identified that a key factor holding back models from vision-language representational compositionality is cross-modal relation alignment. We categorized recent compositional inductive bias approaches into 3 categories: Structural Model, Structural Data, and Structural Training, showing that a previous Structural Training model (IAIS) achieves state-of-the-art performance on Winoground. We then identified a potential key weakness in IAIS, its hard argmax assumption, and developed a soft cross-modal equivalence approach to address it. Having linear algebraically simplified this approach, we arrived at CACR, an auxiliary loss that encourages cross-modal congruence of intra-modal attention. CACR improves on IAIS' performance on Winoground, and even outperforms a UNITER model nearly twice as large.


As computational scaling becomes more widespread, it's necessary to develop compositional inductive biases that do not require complex annotated data or exotic model architectures. Our work illustrates how taking advantage of the transformer's own attentional structure can improve the quality of fine-grained vision-language representations, opening the avenue for large scale approaches to visually-grounded compositionality.

\section{Syntactic MeanPool for Sentence Embedding}
\label{sec:synavg}

Recent work has shown that state-of-the-art vision-language models often fail to represent compositional or relational structure. In simple terms, they are liable to conflate representations of phrases such as "the mug is in the grass" and "the grass is in the mug", resulting in failure to cross-modally distinguish them. Standard practice is to produce embeddings through mean-pooling token embeddings from the final layer. However, in order to capture the compositional structure inherent in language, we instead propose hierarchically averaging token vectors given a syntax parse of the text, a novel Structural Model approach according to the taxonomy of Sec. 4.

\begin{figure}[h]
    \centering
    \includegraphics[width=5in]{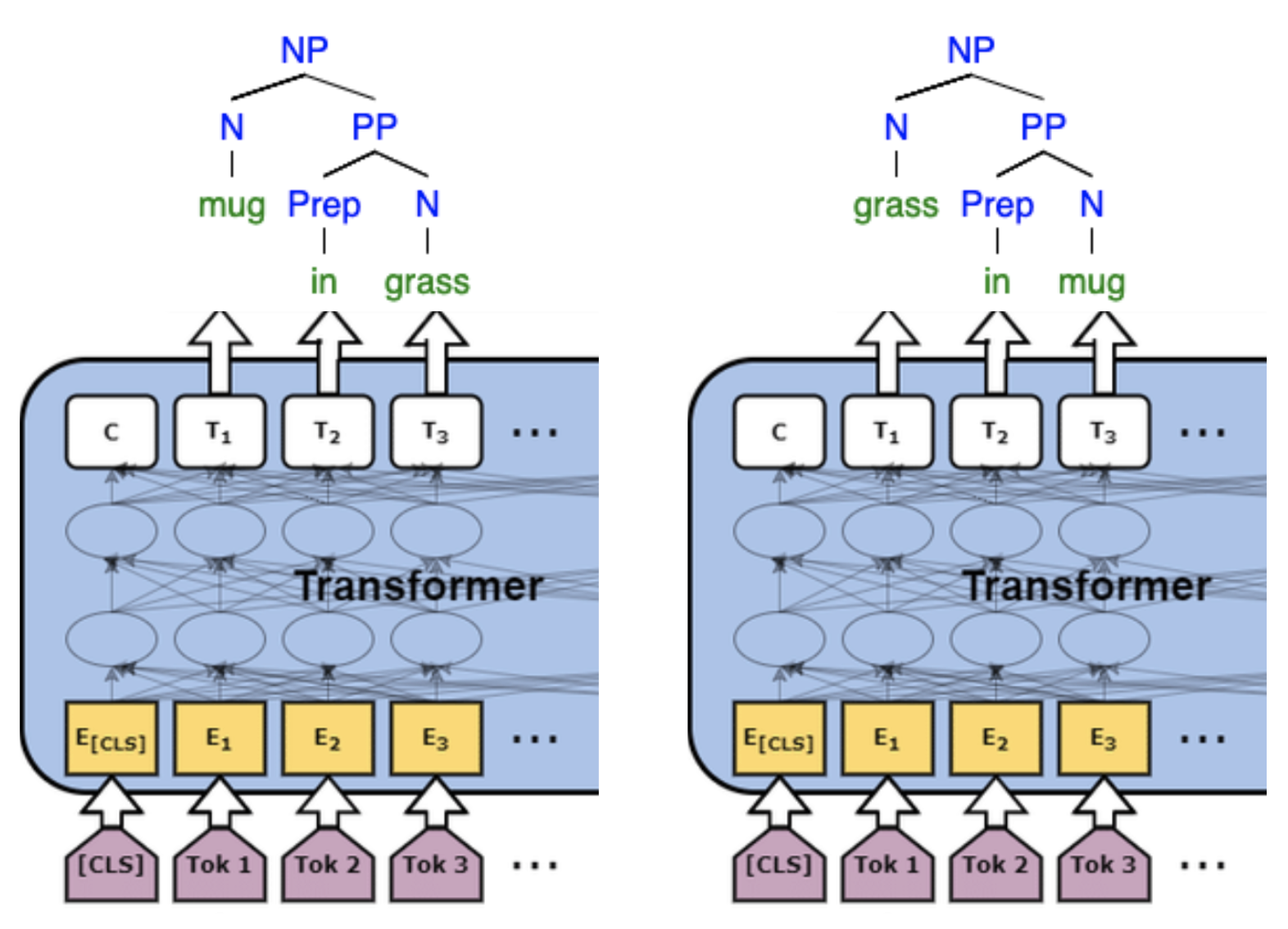}
    \caption{Syntactic MeanPool leverages the non-associativity of the mean operation to inject a compositional inductive bias into the hierarchical pooling of a sentence embedding.}
    \label{fig:synavg}
\end{figure}

We empirically show that our Syntax-guided MeanPool increases the distance between compositionally distinct pairs of sentences compared to MeanPool alone, across multiple embedding models in Tab. \ref{tab:synmeanpool_results}.

\begin{table}[h]
    \centering
    \begin{tabular}{p{6cm}|c c c c c}
         \textbf{Performance} & \textbf{SynMeanPool} & \textbf{MeanPool} & \textbf{CLS Pool} \\
         \hline
         Euc Dist between Winoground captions (BERT) & \textbf{3.21} & 2.14 & 2.57 \\
         Spearman Correlation with STS (BERT) & .302 & \textbf{.476} & .203 \\
         Euc Dist between Winoground captions (CLIP) & \textbf{.382} & .287 & .224 \\
         Winoground Group Score (CLIP) & 6.00 & 7.75 & \textbf{8.25} \\
    \end{tabular}
    \caption{Three sentence embedding pooling strategies, with their influence on Euclidean Distance, Semantic Textual Similarity \cite{cer2017semeval}, and Winoground Score.}
    \label{tab:synmeanpool_results}
\end{table}

Despite increasing Euclidean distance between Winoground captions, intuitively making them more easy to distinguish, why does SynMeanPool not result in increased Winoground score? We suggest that this is likely due to the fact that CLS Pool is the objective that CLIP \cite{radford2021clip} is actually trained for, as evidenced by its Winoground performance being the highest. The obvious next step is to fine-tune CLIP using SynMeanPool to see whether we can successfully inject it with this compositional prior, but we have yet to run this experiment due to implementation issues regarding the PyTorch \cite{paszke2019pytorch} backprop computation graph.



\chapter{Conclusion}
\label{ch:conclusion}

\section{Multimodal Brain}
Multimodal ``Hub \& Spoke" structures have been identified in the human brain that enable embodied semantic cognition in the anterior \& inferior temporal lobe \cite{antonucci08, ralph16}. Through the use of continuous vectorial representations, DL similarly provides a paradigm in which multiple modalities can be aligned to learn joint embeddings of vision and language in the form of VLMs \cite{lu19vilbert}. While VLMs can clearly map simple words to images, it's natural to ask whether they can ground highly compositional utterances as humans do \cite{smolensky2022neurocompositional}.

\section{Vision-Language Psychology}

More broadly, the ability to impose structure upon the world where discrete objects can be related hierarchically is important for human intelligence and likely critical for developing artificial general intelligence. The role of language in imposing these world structures is potentially causal but certainly connected \cite{lu2021pretrained}, a controversial debate in the cognitive sciences \cite{corballis2014recursive}.

This line of work may also comment on debates in linguistics over the innateness of syntax (universal grammar) \cite{yang2004universal}. If structural knowledge can be learned in an unsupervised or self-supervised manner without an innate "syntax module", we contradict the Chomskyan position. However, if we find that structural augmentation methods that require some innate ability to produce hierarchical structures from input are more performant, this could imply that syntactic ability is indeed innate. The architecture of such a syntax module for structural augmentation may then inform or corroborate neuroscience work exploring the cortical basis of universal grammar and its role in multimodal/visual cognition.


In a related work, we present a time constraint based approach to explore how the mind aligns relations between imagery and text. Vision-language relation alignment refers to the ability to determine that an image of a \textit{lightbulb surrounded by plants} is indeed that and not instead `a \textit{lightulb surrounding plants}'. We expose an image to the subject for some amount of time $t$ and then asked them to choose which of two captions—that only differ in relation—matched the image. We find that performing this task takes significantly longer than simple object recognition, requiring 500ms to achieve 75\% and 1000ms to achieve 90\% accuracy. We explore what this might suggest about cognitive architecture, with the lack of a narrow threshold suggesting no explicit module for vision-language relation alignment.

\section{Multimodal Semantics}

Finally, we may seek to draw a more tenuous philosophical parallel between vectorial semantics and contemporary work on type theoretic foundations of mathematics. Past work has sought to treat grammatical types as vectorspaces such that syntactic compositions of types reflect their compositional meaning in the equivalent vectorspace, using algebraic and categorical formalizations \cite{coecke2010discocat}. Concretely, if we have "rabbits hop", our simplified parse tree is [S [NP [rabbits]] [VP [hop]]]. The token "rabbits" is located in the space corresponding to type NP, while "hop" is in the VP space. The syntax rule that composes them (S $\rightarrow$ NP VP) is a type reduction which can be treated as a morphism in the category of vectorspaces, enabling their embeddings to compose according to the sentence's syntax.

\begin{table}[h]
\centering
\begin{tabular}{lllll}
\textbf{Logic} & \textbf{Type Theory} & \textbf{Homotopy} & \textbf{Semantics} & \textbf{Multimodal} \\
\hline
Proposition & Type & Space & Denotation & Text \\
Proof & Term & Point & Referent & Image \\
 &  &  &  & 
\end{tabular}
\end{table}

Using related mathematical machinery, we suggest a parallel between grounded semantic vectorspaces and homotopy types. Homotopy type theory \cite{program2013hott} builds on the Curry-Howard isomorphism by treating propositions (types) as spaces and their proofs (terms of a type) as points in their space. We give HoTT a linguistic interpretation, where spaces correspond to a word's semantic denotation and points in the space are intuitively its grounded referents. Thus the meaning of "rabbit" is captured by the topological structure of the embedding subspace consisting of images of rabbits. Composing these spaces is enabled by the same type theoretic operations that underlie logical composition in HoTT. Denotation spaces may hierarchically nest in broader type universes corresponding to semantic fields or syntactic types. Whether any of this is empirically true or derivable from VLM embedding spaces remains yet to be shown, but such a formalism may elegantly capture grounded meaning in a way historically unattempted in formal semantics.

\subsection{Indian Philosophy}
My interest in multimodal semantic compositionality is not simply a pragmatic concern with what I believe is critical to develop Artificial General Intelligence. Rather, it’s a direct consequence of my ethno-cultural identity. Philologists have noted that the importance of Pythagoras \& Euclid’s mathematics in the Western intellectual tradition is much the same as P{\=a}\d{n}ini’s linguistics in India \cite{staal1965euclid}; the latter developed the first ever generative grammar using a concise system of symbols, functions, and meta-rules. 

The Sanskrit tradition directly contributed to the rise of historical and structural linguistics in the West, with pioneers like Saussure \& Chomsky directly citing the work of Bhart\d{r}hari \& P\=a\d{n}i\d{n}i thousands of years prior \cite{pandey2022indian}. In fact, India witnessed extensive debate on issues of semantic compositionality \cite{saxena2019linguistic} \& perceptual grounding \cite{bronkhorst2011note} among the Ny{\=a}ya, M{\=\i}m{\=a}\d{m}s{\=a}, and Buddhist schools c. 6th-12th centuries AD. In this way, I see my work on the word-world interface as simply continuing the work of my ancestors using the best tools available to me—computers (and a whole lotta GPUs).



\renewcommand*{\bibname}{References}

\addcontentsline{toc}{chapter}{\textbf{References}}

\addcontentsline{toc}{chapter}{APPENDICES} 
\appendix

%


%
\bibliographystyle{plainnat}
\ifthenelse{\boolean{PrintVersion}}{
\cleardoublepage 
}{
\clearpage       
}
\phantomsection  

\bibliography{main}

\end{document}